\documentclass[preprint,12pt]{elsarticle}

\UseRawInputEncoding
\usepackage{amssymb}
\usepackage{graphicx}
\usepackage{caption}
\usepackage{subcaption}
\usepackage{xcolor}
\usepackage{tabularray}
\usepackage{todonotes}
\usepackage{amsmath}
\usepackage{url}
\UseTblrLibrary{booktabs}

\urldef{\mailsa}\path|{grios, pdalbianco, fronchetti, fquiroga, whasperue}@lidi.info.unlp.edu.ar|

\journal{Applied Soft Computing}

\begin{document}

\begin{frontmatter}



\title{Bringing Balance to Hand Shape Classification: Mitigating Data Imbalance Through Generative Models}


\author[lidi,infounlp]{Gaston Gustavo Rios\corref{cor1}}
\ead{grios@lidi.info.unlp.edu.ar}
\author[lidi,infounlp]{Pedro Dal Bianco}
\author[lidi,cic]{Franco Ronchetti}
\author[lidi,cic]{Facundo Quiroga}
\author[lidi,infounlp]{Oscar Stanchi}
\author[lidi,infounlp]{Santiago Ponte Ahón}
\author[lidi]{Waldo Hasperué}

\affiliation[lidi]{organization={Instituto de Investigación en Informática LIDI - Universidad Nacional de La Plata},
            addressline={50 \& 120}, 
            city={La Plata},
            postcode={1900}, 
            state={Buenos Aires},
            country={Argentina}}

\affiliation[cic]{organization={Comisión de Investigaciones Científicas de la Provincia de Buenos Aires (CICPBA)},
            city={La Plata},
            postcode={1900}, 
            state={Buenos Aires},
            country={Argentina}}

\affiliation[infounlp]{organization={Becario Doctoral - Universidad Nacional de La Plata},
            addressline={50 \& 120}, 
            city={La Plata},
            postcode={1900}, 
            state={Buenos Aires},
            country={Argentina}}

\cortext[cor1]{Corresponding author.}

\begin{abstract}
Most sign language handshape datasets are severely limited and unbalanced, posing significant challenges to effective model training. In this paper, we explore the effectiveness of augmenting the training data of a handshape classifier by generating synthetic data. We use an EfficientNet classifier trained on the RWTH German sign language handshape dataset, which is small and heavily unbalanced, applying different strategies to combine generated and real images.
We compare two Generative Adversarial Networks (GAN) architectures for data generation: ReACGAN, which uses label information to condition the data generation process through an auxiliary classifier, and SPADE, which utilizes spatially-adaptive normalization to condition the generation on pose information. ReACGAN allows for the generation of realistic images that align with specific handshape labels, while SPADE focuses on generating images with accurate spatial handshape configurations.
Our proposed techniques improve the current state-of-the-art accuracy on the RWTH dataset by 5\%, addressing the limitations of small and unbalanced datasets. Additionally, our method demonstrates the capability to generalize across different sign language datasets by leveraging pose-based generation trained on the extensive HaGRID dataset. We achieve comparable performance to single-source trained classifiers without the need for retraining the generator.


\end{abstract}


\begin{highlights}
\item GAN-generated datasets improve handshape classification 
 performance on limited and unbalanced data
\item Using generated samples for pre-training and real samples for fine-tuning is key to boosting model performance.
\item Dataset balance through generative models boosts per-class accuracy by up to 100\% in several cases.
\item Models pre-trained with generated samples achieve an earlier convergence.
\item Our models set a new state-of-the-art for the RWTH handshape dataset.
\end{highlights}


\end{frontmatter}





\section{Introduction}

In recent years, the performance of deep learning models has improved significantly. However, this progress is closely related to the availability of large high-quality datasets, which are often difficult and expensive to create \cite{Bragg2021}. The challenge is especially pronounced in sign language recognition, where data scarcity and imbalance \cite{Rezvani2023,Yu2024} are prevalent. Many sign language datasets suffer from a lack of diversity and volume, as they require the participation of signers for accurate data collection and labeling. This results in small, unbalanced, and low-quality datasets \cite{Bragg2019}, limiting the performance of the models trained on them \cite{Koller2016,DalBianco2022,Nunez2023}.

Since data collection is difficult, sign language data is generally obtained from real-world sources. Due to the natural distribution of signs and words within a language, and the fact that many data sources focus on a limited range of themes, most sign language datasets tend to be naturally unbalanced \cite{Li2020,Koller2016,DalBianco2022}. Moreover, the creation of new sign language datasets is further hindered by the fact that sign languages are not mutually intelligible, necessitating the development of separate datasets for each language \cite{Bragg2021}. As a consequence, communities with fewer resources are disproportionately affected, with even high-resource communities facing significant challenges due to the limited scope and quality of available datasets.

Synthetic training data generation has proven to be effective in improving model training in limited and unbalanced datasets, leading to faster and more stable convergence \cite{Mostofi2024,Gaggiotti2021,Sampath2021,Xia2023}. However, the generated images often lack realism, introducing noise that can degrade the training process. Furthermore, label-based generation struggles with generalization across domains, as it requires a specific generative model for each sign language \cite{Gaggiotti2021}. Despite significant advancements in multi-domain generators \cite{Betker2023,Rombach2021}, these models still fail to produce accurate and realistic images for specialized domains such as sign language handshapes. Thus, there remains a critical need for a general-purpose handshape generator that can operate effectively across multiple sign languages.

\subsection{Proposed approach}

In this article, we propose using generated data to improve the classification of handshapes on datasets with unbalanced and limited data.

To augment the datasets, we propose the Generative Adversarial Networks (GAN) architectures conditioned on labels and pose. Rebooted Auxiliary Classifier GAN (ReACGAN) uses labels to calculate the data cross-entropy (D2D-CE) loss which is used with the adversarial loss to train the model. In contrast,  SPatially-Adaptive
(DE)normalization (SPADE) replaces Conditional Batch Normalization as the conditional normalization method for our second model which we refer to simply as SPADE and receives pose data as part of its input. Given that pose information can be extracted from any sign language, we can exploit this domain superposition to create a generator capable of generating hand shapes from any sign language. With this in mind, we can easily extend the proposed methods to other datasets.

\begin{figure}[ht!]
    \centering
    \begin{subfigure}{0.33\textwidth}
        \centering
        \includegraphics[height=\hsize]{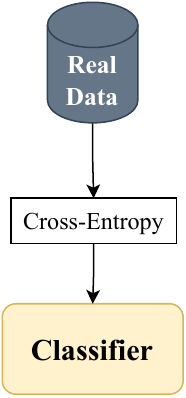}
        \caption{Train only with real data}
        \label{fig:real-train}
    \end{subfigure}
    \hfill
    \begin{subfigure}{0.33\textwidth}
        \centering
        \includegraphics[height=\hsize]{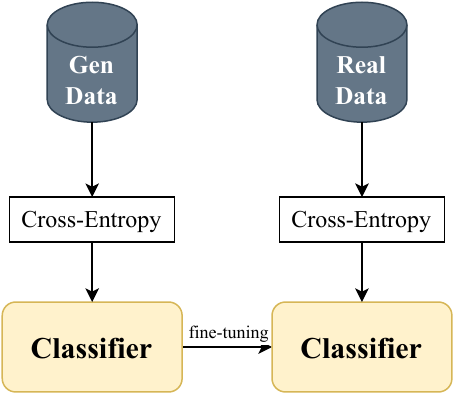}
        \caption{Pretrain with generated data}
        \label{fig:pregen-train}
    \end{subfigure}
    \hfill
    \begin{subfigure}{0.33\textwidth}
        \centering
        \includegraphics[height=\hsize]{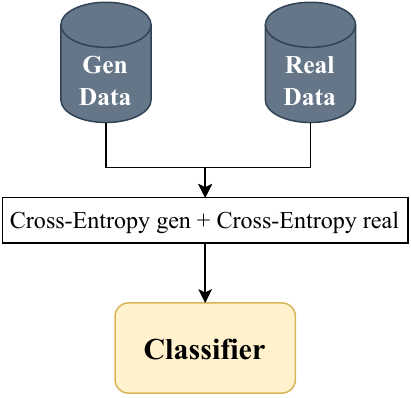}
        \caption{Regularize training using generated data}
        \label{fig:regen-train}
    \end{subfigure}
    \hfill
    \begin{subfigure}{0.33\textwidth}
        \centering
        \includegraphics[height=\hsize]{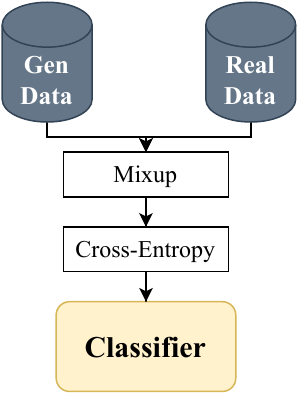}
        \caption{Use generated data with mixup}
        \label{fig:mixgen-train}
    \end{subfigure}
    \caption{Diagram (a) shows the regular training approach of our classifier model. For our other methods, a generator model is fitted with real data to create newly generated data samples. 
    }
    \label{fig:methods}
\end{figure}

We compare several approaches to take advantage of the generated data (Figure \ref{fig:methods}).

\begin{itemize}
    \item REAL: pre-training on ImageNet, and fine-tuning with real data. Used as a baseline.
    \item PRETRAIN: pre-training with generated data, and fine-tuning on real data
    \item REGULARIZER: Training with both generated and real data, using the generated data as a regularizer
    \item MIXUP: Training with both generated and real data, using mixup to combine them.
\end{itemize}

\subsection{Contributions}

Our work introduces several key contributions that advance the field of sign language handshape classification, particularly in the context of unbalanced and limited datasets:

\begin{itemize}
\item Improved Classification and Per-Class Accuracy: We demonstrate that augmenting the training dataset with GAN-generated samples can significantly improve the accuracy of handshape classification. Specifically, our method achieves a 5\% improvement over the state-of-the-art on the RWTH German sign language dataset. By generating a balanced dataset with GANs, we were able to correctly classify underrepresented classes that could not be accurately classified when training only with real data. This dual benefit addresses both the general performance and the specific challenge of class imbalance.
\item Effective pre-training Strategy: We conducted a comprehensive comparison of different training strategies using a combination of generated and real data. Our findings show that pre-training with GAN-generated samples, followed by fine-tuning on real data, yields superior performance compared to alternative approaches.
\item Accelerated Convergence: We observe that models pre-trained with GAN-generated data converge more rapidly during training. This faster convergence not only reduces computational costs but also enhances the efficiency of the training process, making it more feasible to deploy high-performing models in real-world applications.
\item Generalization Across Datasets: We explore the use of both class-based and pose-based data generation strategies. While both methods enhance model performance, pose-based generation proves particularly effective in enabling the generalization of the model to multiple handshape datasets from different sign languages. This contribution highlights the versatility of our approach in addressing the diversity of sign language datasets.
\end{itemize}




\section{Related Work}

Class imbalance is a commonly found problem in machine learning tasks. This problem is mainly been approached with data-level methods and algorithm-level methods \cite{Zhu2022}. Data-level methods modify the data distribution either by adding, removing, or applying data augmentation over the original dataset. In contrast, algorithm-level methods create new algorithms and loss functions that favor minority classes.

Generated synthetic data can improve training and increase the efficiency of data for models with limited and unbalanced data by introducing new instances \cite{Tremblay2018, Harkirat2020, Kortylewski2018, Kim2019, Jiang2009, Ibrahim2012, Fowley2022, Binod2019}. Generated data can be classified according to their sources, each creating different types of new data samples via image transformation, simulation, or neural network inference.

\subsection{Image transformation}
Data augmentation via image transformation has been used to prevent overfitting in deep learning algorithms. It can be introduced to any model training with little computational cost \cite{Krizhevsky2012}. 
This method works by applying a randomly selected set of transformations to each input image. These transformations include noise injection \cite{Moreno-Barea2018}, random erasing \cite{Zhong2020}, RGB channels alterations, and geometric transformations such as translation, rotation, or reflection \cite{Krizhevsky2012}. By introducing these transformations, it is possible to create new synthetic data that can be used to train a model. Because this data is created by applying transformations, it is limited by its source. In addition, the transformations can change the intended label of the sample when it is too strong. 

Image transformation has been used to train models on sign language datasets \cite{Koller2020,Kim2019}. It has been shown to improve the performance of models trained with this method by almost 3\% on the RWTH-PHOENIX-Weather (RWTH) handshapes dataset \cite{Koller2016} when the transformation is not too aggressive \cite{cornejo2019}.

\subsection{Simulation}

Data generated artificially using a simulator can provide an unlimited amount of new data samples under predefined conditions. The limitation of this method is that each sample or at least each element present in each sample must be created individually. This makes this method time-consuming, which limits its usability. However, this method has been proven useful to improve the training of models by training with synthetic and real data \cite{Shrivastava2016, Tremblay2018, Kortylewski2018, Zimmermann2017, Ibrahim2012}.

\subsection{Neural network inference}

In the last few years, generative models have shown great improvements in the quality of synthetic images\cite{Baltatzis2024}. The most successful models, such as Generative Adversarial Networks (GAN), Variational Auto Encoders (VAE), and Diffusion Models, can generate realistic new images without memorizing the data in the training set \cite{Vaishnavh2019,Arora2017}. GAN works by jointly training a discriminator and a generator, where the generator minimizes the distance between the generated and real data so that the discriminator cannot discern them. With these models, we can generate an arbitrary number of new data samples that do not rely on prior assumptions about the data distribution. However, these images may show artifacts that end up adding noise when training new models with them. Furthermore, mode collapse can affect the variation of the generated images, resulting in a limited number of unique images. Nonetheless, the new images created by these models can be used to augment the training data when dealing with limited \cite{Bowles2018,Maayan2018} and unbalanced \cite{Sampath2021} data. Smart sampling techniques can improve the performance of models trained with generated data by discarding lower-quality samples \cite{Binod2019} and keeping only the top-K best samples.

Other well-known models that can generate high-quality images are diffusion models. These models compete in quality and diversity with GAN models, even beating them on occasions in realism \cite{Dhariwal2021,MullerFranzes2023} but with inferior inference speed \cite{Vahdat2022}. 
Diffusion models are trained using two processes: forward diffusion and parametrized reverse diffusion. The generative process then consists of many denoising steps that generate a realistic image from noise \cite{Cao2023}. This can make the inference process to generate new samples slow as for each new image there may be thousands of denoising steps. 


\subsubsection{ReACGAN}
ReACGAN \cite{Kang2021} was proposed as an improvement on the methods used by Auxiliary Classifier GAN (ACGAN) \cite{Odena2016}. ACGAN \cite{Odena2016} uses conditional information during training by jointly using the classification and source losses which increased its performance over the regular GAN. However, ACGAN has been shown to have unstable training when the number of classes increases and to collapse to a small amount of easily classifiable generated data. These problems are addressed by ReACGAN by projecting input vectors onto a unit hypersphere and using data-to-data class comparisons at each mini-batch. ReACGAN achieves state-of-the-art results and has comparable performance to many diffusion models \cite{Kang2021}.

\subsubsection{SPADE}
SPADE \cite{Park2019} is a conditional GAN originally intended to use a segmentation layer to condition the generation of new synthetic data. It introduces a new normalization method similar to the Conditional Batch Normalization module that allows the usage of 2D data by employing convolutions. This allows us to condition the model on the 2D representation of the joints and bones of the hands.



\section{Methodology}

In this section we describe our proposed generative model-based data augmentation methods for handshape classification. By using generated data we aim to improve domain generalization \cite{Kaiyang2021} and increase model robustness against out-of-distribution data. This approach can be either single-source or multi-source, depending on whether the generator is trained on the same dataset as the classifier or on multiple related domains. In the single-source case, the generator can capture the distribution of the original dataset and generate new samples that have similar properties to the real data. In the multi-source case, the generator can learn to generate images that have a broader range of variability and diversity by leveraging information from multiple related domains. In this paper, we train our classifier on RWTH using both single-source and multi-source methods. For our multi-source training, we first train a pose-conditioned generator on HaGRID, taking advantage of the higher size and variability of the hands to create a generic hand generator. Then, this generator is used with RWTH poses to create a synthetic dataset that is used jointly with real images to train the classifier in a multi-source way.

Augmenting the training data with new images obtained from a trained generator is a powerful technique to improve generalization and reduce overfitting when labeled data is limited or expensive to obtain. This approach enables the creation of a more varied distribution of data and can effectively increase the size of the dataset without requiring additional labeling effort. Data can be generated accounting for class balance to lessen the impact of the original data class imbalance.

In the case of sign language, we can train the generator on multiple languages or hand gestures that share the domain of hand poses to increase the variety and quality of the generated images. This can be thought of as multi-task learning \cite{ZhangYu2017}, where multiple similar tasks are learned to improve the training on a new related task. The independent generator facilitates the knowledge transfer between these domains by learning to generate images for each one simultaneously. Conditioning the generator on poses then enables the creation of images belonging to specific classes of the target with the added semantics of each source domain. On the other hand, using labels as input can result in more specific generated samples, as the network can focus on learning the features that are most relevant to each class. However, it does not incorporate the additional knowledge gained from other sources.



We evaluate three alternative training methods using synthetic data. These methods are: training using generated data and real data, using generated data as regularization, and using generated and real data with mixup. The different methods can be seen in Figure \ref{fig:methods}, the loss used by each method is displayed in Table \ref{tab:methodssummary}. We also compare the usage of raw generated data and filtered generated data, where we filter out the worst samples of the synthetic dataset, in a similar way to the robust learning method of source weighting \cite{Konstantinov2019}. To see the impact of using generated data in the training of small datasets we sub-sampled RWTH and HaGRID to obtain several smaller subsets. We then ran the experiments using these reduced datasets.

\subsection{Formal description of the training approaches}

In the following subsection, we will proceed with the formal description of each training method. Refer to Table \ref{tab:symbols} for the definition of the symbols used in this section.

\begin{table}[ht!]
	\centering
	\footnotesize
	\centerline{\begin{tblr}{
        colspec={cc},
    	row{1}={font=\bfseries},
    	row{even}={bg=gray!10},
	}
    	\textbf{Symbol} & Definition \\
    	\toprule
        \(X\)      	& Classifier input space (image) \\
    	\(Y\)     	& Classifier output space (label) \\
    	\(c\)		& Classifier \(c:X\to Y\) \\
    	\(S_r\)     	& Real samples \\
    	\(\mathcal{L}\)	& Classifier loss function \\
        \(\epsilon\)	& Training epoch  \\
    	\(s\)     	& Image quality score \\        
        & \\        
        \bottomrule
    \end{tblr}
    \begin{tblr}{
        colspec={cc},
    	row{1}={font=\bfseries},
    	row{even}={bg=gray!10},
	}
    	\textbf{Symbol} & Definition \\
    	\toprule
    	\(Z\)     		& Generator input space (noise and condition) \\
        \(X’\)		& Generator output space (image and label) \\
    	\(g\)     	& Generator \(g:Z\to X’\) \\
    	\(S_g\)	& Generated samples \\
    	\(\mathcal{L}_{reg}\)     	& Classifier regularized loss function \\
        \(\sigma\)	& Generated data weight during training \\
        \(\alpha\)     	& \(\sigma\) starting value during regularization \\
        \(\beta\)	& \(\sigma\) change rate during regularization \\
        \bottomrule
    \end{tblr}}
    \caption{Definitions of the key symbols and variables used in the proposed methods.}
	\label{tab:symbols}
\end{table}

\begin{table}[ht!]
	\centering
	\footnotesize
	\centerline{\begin{tblr}{
        colspec={ccc},
    	row{1}={font=\bfseries},
    	row{even}={bg=gray!10},
	}
    	Method & Pretrain loss & Train loss \\
    	\toprule
    	real		& -		& \(\mathcal{L}(S_r)\) \\
    	pretrain	& \(\mathcal{L}(S_g)\)	& \(\mathcal{L}(S_r)\) \\
    	regularization	& -		& \(\mathcal{L}(S_r)+\sigma(x,\alpha,\beta)\mathcal{L}(S_g)\)
 \\
    	mixup		& -		& \begin{tabular}[x]{@{}c@{}}\(\mathcal{L}(\text{mixup}(S_r,S_r))\;\;\;\text{if}\;U(0,1)<\sigma(x,\alpha,\beta)\)\\\(\mathcal{L}(\text{mixup}(S_r,S_g))\;\;\;\text{if}\;U(0,1)>\sigma(x,\alpha,\beta)\)\end{tabular}
 \\
           \bottomrule
    \end{tblr}
    }
    \caption{Summary of loss functions for different training methods.}
	\label{tab:methodssummary}
\end{table}

\subsubsection{pre-training using generated data}
In this method, we first train the model \(c\) using \(S_g\) and then fine-tune it using \(S_r\). \(S_g\) provides generalization to the model trained with it as it creates new samples not contained in \(S_r\). This method exploits the continuous property of \(S_g\) to avoid falling in a local minima of the loss function and overfitting. \(g\) can generate interpolations between the different classes smoothing the boundaries between them. Fine-tuning the classifier \(c\) with \(S_r\) further increases the performance on the given task as the data in \(S_g\) can contain artifacts or imperfections.

\subsubsection{Regularization using generated data}


We also explored training \(c\) using real data while incorporating generated data as a regularization term. By applying a weighted regularization term in the loss we obtain a regularized loss \(\mathcal{L}_{reg}\) which includes the cross-entropy functions for both the real dataset \(S_r\) and generated dataset \(S_g\). To improve the model's generalization when trained with generated data without compromising its ability to learn from the original data, we introduce a dynamic weighting parameter \(\sigma\). This parameter changes its value as the training advances depending on the training epoch \(\epsilon\), a starting value \(\alpha\), and a change rate \(\beta\).
\begin{equation}
\mathcal{L}_{reg}=\mathcal{L}(S_r)+\sigma(\epsilon,\alpha,\beta)\mathcal{L}(S_g)
\end{equation}
This method seeks to use the generalization provided by the generated data to prevent falling on local minima in \(\mathcal{L}_{reg}\) by guiding the training with the regularization term. The parameter \(\sigma\) is calculated in two different ways, one for increasing its value during training \(\sigma_{\uparrow}\) and one for decreasing it \(\sigma_{\downarrow}\).

\begin{equation}
\label{eq:inc}
\sigma_{\uparrow}(\epsilon,\alpha,\beta)=\alpha+(1-\alpha)(1-e^{-\beta \epsilon})
\end{equation}
\begin{equation}
\label{eq:dec}
\sigma_{\downarrow}(\epsilon,\alpha,\beta)=\alpha e^{-\beta \epsilon}
\end{equation}

\subsubsection{Real and generated mixup}


Synthetic mixup creates new virtual training samples by combining two inputs. These inputs are drawn randomly from \(S_r\) and \(S_g\) as pairs \((x, y)\) and \((x’, y’)\). \(\lambda \in [0,1]\) is a random value that determines the weight of each of the 2 inputs.
\begin{equation}
\tilde{x}=\lambda x’+(1-\lambda)x_r
\end{equation}
\begin{equation}
\tilde{y}=\lambda y’+(1-\lambda)y_r
\end{equation}

Data augmentation using mixup has been shown to increase the robustness and generalization of models that are trained with it \cite{ZhangHongyi2017, Hataya2019}. This increases robustness by minimizing an upper bound on adversarial loss \cite{Zhang2020}. The usage of generated data further uses this interpolation by creating an almost unlimited amount of interpolations. We used \(\sigma \in [0,1]\) to change the impact of the generated data during training. In this scenario, \(\sigma\) changes the probability of using synthetic mixup over regular mixup with two real samples.

\subsubsection{Filtered generated data}

To prevent the noise present in the generated data from degrading our model performance, we sub-sample the generated data. We reduce the noise present in \(S_g\) and reduce the impact of the worst samples by filtering them when the quality is low. To this end, we use class conditional probabilities \cite{Binod2019} to rank and score each sample. 

We first train a regular classifier using the EfficientNet v2 \cite{Mingxing2021} architecture on real data. This classifier is then used to score each generated image individually obtaining a pair \((x’_i, s_i)\) where \(x’_i\) is the \(i^{th}\) generated image and \(s_i\) its associated score, which is the probability of the image of belonging to its correct class. Using the scores assigned to each image, we rank them and select the top-K highest-scoring samples for training the final classifier. By removing the lower-ranking samples, we ensure that the training data consists of the images that most closely resemble their respective classes.

\section{Experiment settings}

\subsection{Datasets}

\begin{figure}[ht!]
    \centering
    \includegraphics[width=\hsize]{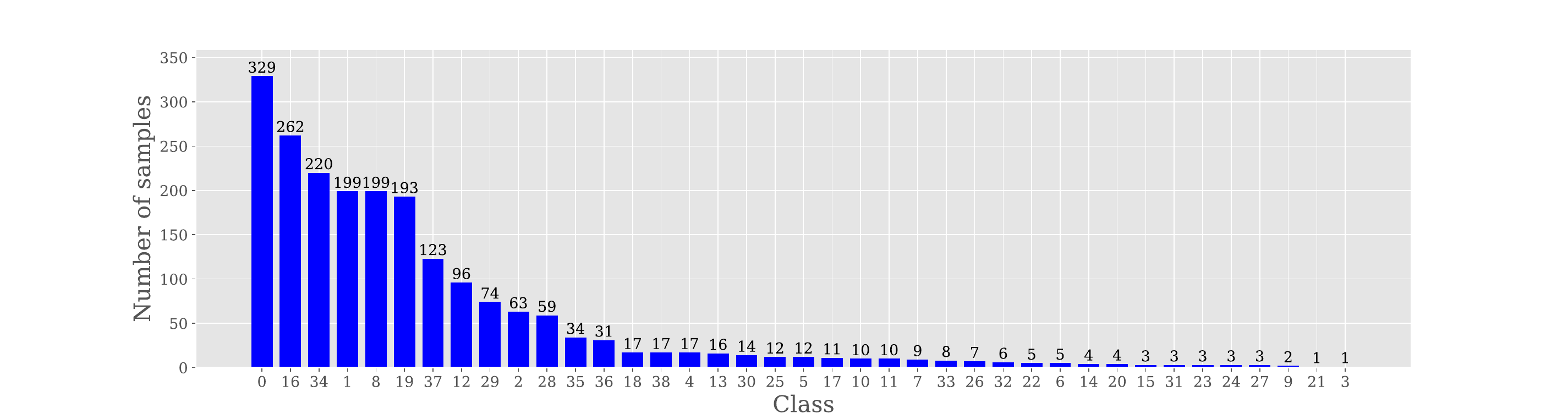}
    \caption{Count of training images belonging to the 39 hand shapes of RWTH. Each hand shape is assigned a number as its class label. The dataset is highly imbalanced, with only 7 out of 39 classes having more than 100 samples and 16 classes having less than 10 samples.}
    \label{fig:classunbalancerwth}
\end{figure}

RWTH-PHOENIX-Weather (RWTH) \cite{Koller2016} is composed of 3359 labeled images of signs captured from the German public TV-station PHOENIX. After doing pose detection over these images, the total was reduced to 2098. The dataset contains a total of 39 different hand shapes after this reduction. The signs belong to the German sign language. All images were cropped centered on the signers and resized to a size of 132x92. The dataset is highly unbalanced and contains a large intra-class variance and similarity between different classes. Figure \ref{fig:classunbalancerwth} shows the count of samples of each class, where the 10 most numerous classes contain more than 80\% of the images. The interpreters always wear black clothes over a white background. We used the 1 million weakly labeled images also provided in the same dataset to train the generator.

HaGRID \cite{Kapitanov2022} was created for static hand gesture classification and detection. Although it is not a sign language dataset, the domains are similar enough to analyze the effectiveness of our techniques. Furthermore, for some techniques this distinction is not relevant, since we can use the dataset simply as a source of hand images in different poses. HaGRID consists of 552,992 FullHD (1920 × 1080) RGB images of 18 hand gesture classes and a “no gesture” class. These images were collected, validated, filtered, and annotated using two different crowdsourcing platforms. There are a total of 34730 unique persons, each with a different scene. HaGRID shows high diversity between each person, lighting, and background. We decided to crop the hands because the 64x64 resolution used in this paper is not enough to accurately distinguish gestures. The dataset also provides the 2D coordinates of 21 keypoints for each hand, which represent the locations of the fingers and palm.


\subsection{Data preprocessing}

For datasets that did not include pose information, we extracted the hand poses using OpenPose \cite{Cao2019}. This resulted in a total of 21 keypoints per hand, but since we work with single-hand signs, we only use the hand with higher confidence for each image. We removed any samples for which we could not extract any pose. 

We then cropped each image to 64x64 pixels centered on the hands, normalized the pixel values, and randomly flipped each image to augment the dataset. We separated some of the samples of each dataset to use as our held-out test set for our classifier. The remaining data was used for training and validation of both classifier and GAN models. For the reduced datasets experiments, we decreased the amount of available data of the training set in a stratified way. This same reduced training set was used to train the GAN models and the classifier.

\begin{figure}[ht!]
    \centering
    \begin{subfigure}{0.3\textwidth}
        \includegraphics[width=\hsize]{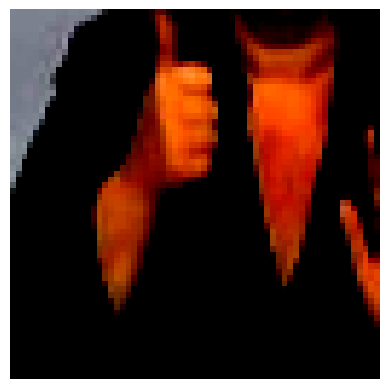}
        \caption{Original}
        \label{fig:rwthok}
    \end{subfigure}
    \hfill
    \begin{subfigure}{0.3\textwidth}
        \includegraphics[width=\hsize]{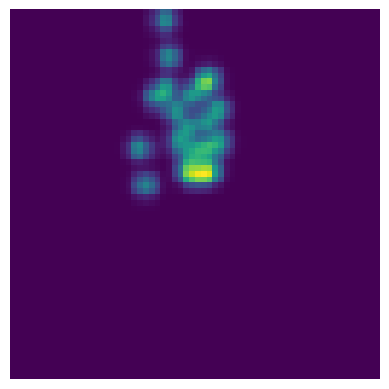}
        \caption{Joints}
        \label{fig:rwthokkeypoints}
    \end{subfigure}
    \hfill
    \begin{subfigure}{0.3\textwidth}
        \includegraphics[width=\hsize]{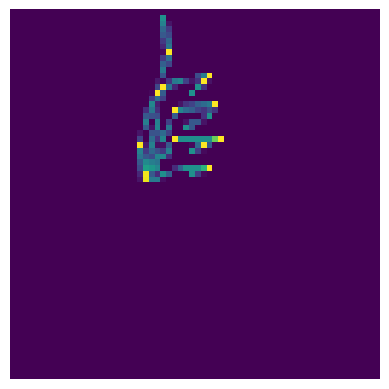}
        \caption{Bones}
        \label{fig:rwthokpose}
    \end{subfigure}
    \caption{Visualization of the original image, its joints, and bones. Each keypoint consists of a channel containing a multivariate normal distribution centered on the keypoint location. Each bone consists of a channel containing a line that joins two anatomically adjacent keypoints.}
    \label{fig:keypoints}
\end{figure}

To incorporate the pose information into our generative models, we developed two different techniques. The first approach, shown in Figure \ref{fig:rwthokkeypoints}, involves creating a multivariate normal distribution with mean at each keypoint, with a small covariance matrix. Each of these distributions is then separated into an individual channel to ensure that no information is lost if two keypoints overlap. The second technique, shown in Figure \ref{fig:rwthokpose}, involves drawing a line at each hand joint. As with the first approach, each line is assigned an individual channel.

\subsection{Generative models}

\begin{figure}[ht!]
    \centering
    \includegraphics[width=\hsize]{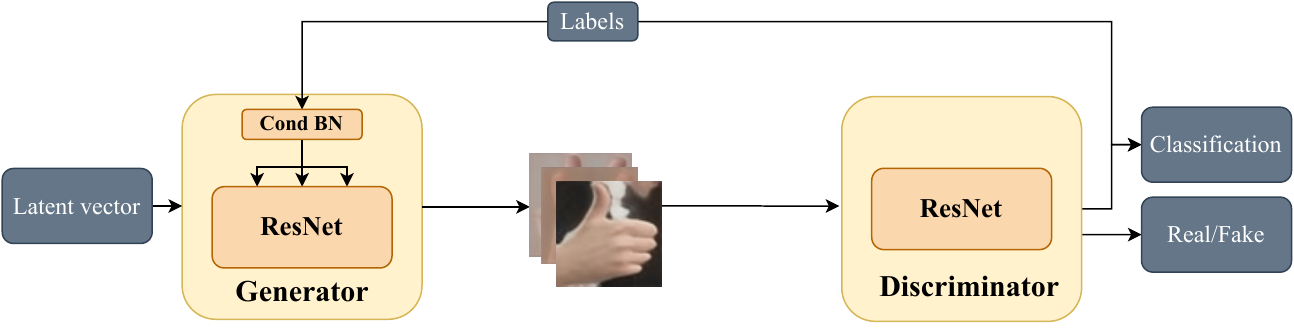}
    \caption{Diagram depicting the ReACGAN model. The generator takes as input a latent vector, sampled from a Gaussian distribution, and a label. The discriminator takes as input a generated or real image. The discriminator then uses its outputs to calculate the \textit{Data-to-Data Cross-Entropy} (D2D-CE) and adversarial losses.}
    \label{fig:labeldiagram}
\end{figure}

\begin{figure}[ht!]
    \centering
    \includegraphics[width=\hsize]{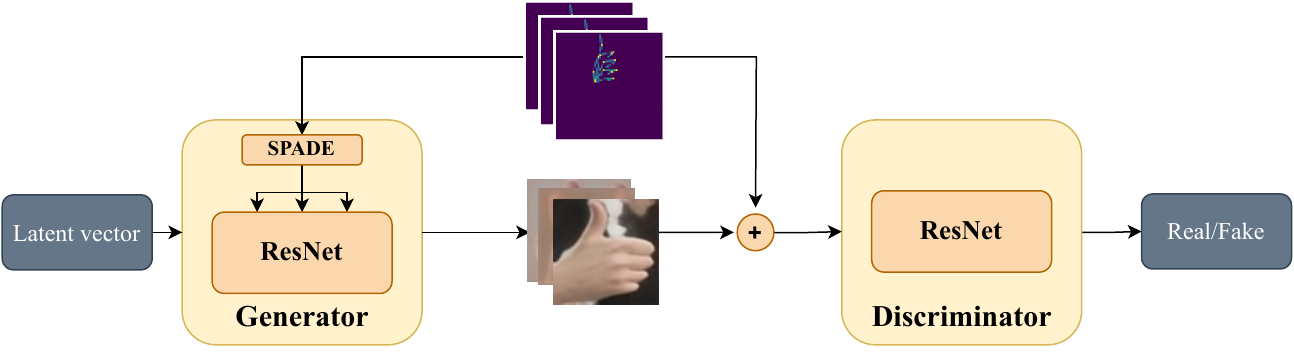}
    \caption{Diagram depicting the SPADE model. The generator takes as input a latent vector, sampled from a Gaussian distribution, and a pose with $c$ channels of the same shape as the output image. The discriminator takes as input the concatenation of the generated or real images and their respective poses. Then, the output of the discriminator is used to calculate the adversarial loss.}
    \label{fig:posediagram}
\end{figure}

To compare the effectiveness of generating images from different sources, we trained multiple generator architectures and compared the performance of the resulting classifiers trained on the generated data. Specifically, we used Generative Adversarial Networks (GAN) conditioned on labels, hand joints, and hand bones. To condition on the label, we used an auxiliary classifier\cite{Kang2021}, while SPADE \cite{Park2019} layers are used to condition on the hand joints and bones. The diagram of each model can be seen in Figures \ref{fig:labeldiagram} and \ref{fig:posediagram}.
We chose GANs over Diffusion Models given their faster inference speed and similar performance, which allows for the generation of large synthetic image datasets in a short amount of time. This capability can be beneficial for dynamically reducing overfitting, similar to the benefits of active learning \cite{Pengzhen2020}.

\section{Experiments and results}

\subsection{Hanshape generation}

\begin{table}[ht!]
    \centering
    \footnotesize
    \begin{tblr}{
        colspec={llllll},
        row{1}={font=\bfseries},
        column{1}={font=\itshape},
        row{even}={bg=gray!10},
    }
        \textbf{RWTH} & FID($\downarrow$) & IS($\uparrow$) & Coverage($\uparrow$) & Density($\uparrow$) & Human($\uparrow$) \\
        \toprule
        ReACGAN $\lambda_{cond}=0.5$ & 45.45          & 2.21          & 0.52          & 0.33         & - \\
        ReACGAN $\lambda_{cond}=1$   & \textbf{45.19} & 2.11          & \textbf{0.60} & \textbf{0.48} & 2.74 \\
        SPADE-keypoints              & 51.96          & 2.24          & 0.43          & 0.30         & - \\
        SPADE-bones                  & 51.05          & \textbf{2.25} & 0.38          & 0.23         & 2.32 \\       
        \bottomrule
    \end{tblr}
    \caption{Comparison of GAN models performance on RWTH dataset. The table shows the results of evaluating multiple GAN models using the Fréchet Inception Distance (FID), Inception Score (IS), Coverage, and Density metrics. We also show a qualitative metric measured using human participants. This metric takes values from 1 to 5, averaging the realism value, in that scale, awarded to each image by the participants. As a point of reference, on real images, this metric average value is 4.4.}
    \label{tab:RWTH-generators}
\end{table}

\begin{table}[ht!]
    \centering
    \footnotesize
    \begin{tblr}{
        colspec={llllll},
        row{1}={font=\bfseries},
        column{1}={font=\itshape},
        row{even}={bg=gray!10},
    }
        \textbf{HaGRID} & FID($\downarrow$) & IS($\uparrow$) & Coverage($\uparrow$) & Density($\uparrow$) & Human($\uparrow$) \\
        \toprule
        ReACGAN        & \textbf{13.9}  & 3.62 & \textbf{0.88} & \textbf{0.80} & \textbf{3.97} \\
        SPADE-bones  & 33.21 & \textbf{3.88} & 0.65 & 0.70 & 1.81 \\
        \bottomrule
    \end{tblr}
    \caption{Comparison of GAN models performance on HaGRID dataset. The table shows the results of evaluating multiple GAN models using the Fréchet Inception Distance (FID), Inception Score (IS), Coverage, and Density metrics. We also show a qualitative metric measured using human participants. This metric takes values from 1 to 5, averaging the realism value, in that scale, awarded to each image by the participants. As a point of reference, on real images, this metric average value is 4.36.}
    \label{tab:HaGRID-generators}
\end{table}

\begin{figure}[ht!]
    \centering
    \begin{subfigure}{0.3\textwidth}
        \includegraphics[width=\hsize]{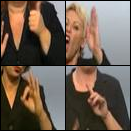}
        \caption{RWTH real}
        \label{fig:rwthoriginal}
    \end{subfigure}
    \hfill
    \begin{subfigure}{0.3\textwidth}
        \includegraphics[width=\hsize]{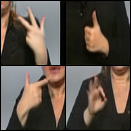}
        \caption{RWTH ReACGAN}
        \label{fig:rwthgeneratedreacganimages}
    \end{subfigure}
    \hfill
    \begin{subfigure}{0.3\textwidth}
        \includegraphics[width=\hsize]{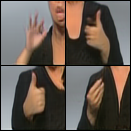}
        \caption{RWTH SPADE}
        \label{fig:rwthgeneratedspadeimages}
    \end{subfigure}
    \hfill
    \begin{subfigure}{0.3\textwidth}
        \includegraphics[width=\hsize]{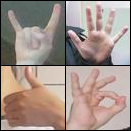}
        \caption{HaGRID real}
        \label{fig:HaGRIDoriginal}
    \end{subfigure}
    \hfill
    \begin{subfigure}{0.3\textwidth}
        \includegraphics[width=\hsize]{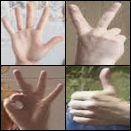}
        \caption{HaGRID ReACGAN}
        \label{fig:HaGRIDgeneratedreacganimages}
    \end{subfigure}
    \hfill
    \begin{subfigure}{0.3\textwidth}
        \includegraphics[width=\hsize]{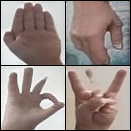}
        \caption{HaGRID SPADE}
        \label{fig:HaGRIDgeneratedspadeimages}
    \end{subfigure}
    \caption{Real and generated samples of RWTH and HaGRID. Generated images were created using the models ReACGAN and SPADE conditioned by label or pose respectively.}
    \label{fig:generatedresults}
\end{figure}

To ensure consistency in our tests we employed the same backbone architecture for all GAN models and applied Spectral Normalization to stabilize training. As our baseline, we use the ReACGAN architecture which uses a residual network backbone. We evaluated our models on 64x64 images of RWTH and HaGRID and conditioned our models using labels with ReACGAN and pose with SPADE. To condition the models on labels Conditional Batch Normalization modules are included in the generator and we use data-to-data cross-entropy for the discriminator. Alternatively, SPADE modules are used to condition the generator with keypoints, then the keypoints are concatenated to the input of the discriminator. We used hinge loss in all cases. We decided to use a high conditional loss to improve the model's capacity to generate images of the correct label. A high conditional loss is necessary to generate images that correctly depict their corresponding labels, this is necessary to reduce the noisiness of the data and prevent a degrading of the classifier when using this synthetic data. Furthermore, increasing the weight of the conditional loss gave a slight improvement to the metrics of the generator model with RWTH. There was no clear difference in the performance of the models conditioned on joints or bones. Therefore, when training HaGRID, we decided to train it using the bones of the hand.

We measured the performance of each model with Fréchet Inception Distance (FID) \cite{Heusel2017} and Inception Score (IS) \cite{Salimans2016}. We pre-calculated the static files of FID using a separate validation set composed of images extracted from the training set. Due to the limits of metrics like IS and FID to measure fidelity and diversity, we also use Density and Coverage \cite{Naeem2020}. This way we can get the degree of resemblance of the real and generated images, and the coverage of the variability of the real samples. In addition, as a qualitative metric we used 11 human participants that assigned each image a score ranging from 1 to 5 indicating the realism of generated images. Each participant was given 10 images for each generated dataset resulting in a total of 110 ranked images per dataset. We also included real images in the forms to compare the scores of real and generated datasets.

Tables \ref{tab:RWTH-generators} and \ref{tab:HaGRID-generators} display the performance of the different GAN models trained on RWTH and HaGRID. ReACGAN showed a consistently better FID, Coverage, Density and Human score. However, SPADE achieved a better IS with both datasets. In HaGRID, ReACGAN achieved more than double the Human score and less than half of the FID of SPADE. This indicates a decrease in the realism of the generated images when using SPADE in comparison with ReACGAN. Figure \ref{fig:generatedresults} shows some of the generated samples with the best FID. 

\subsection{Hanshape classification}

For our classifier we used EfficientNet v2 M \cite{Mingxing2021} due to its excellent performance and fast training. The M version is a scaled-up version of EfficientNet v2 S, with 54M parameters, which further enhances the model’s performance. We optimized using Adam with a learning rate of 1e-4, betas of 0.0 and 0.999, and an epsilon value of 1e-6. Weights are initialized by doing transfer learning from ImageNet unless otherwise indicated. We fine-tuned a decay and growth factor for the generated part of the regularization and mixup with generated data methods. We used the same training samples that were used to train the generator to train the classifier. For each dataset, the model was trained using all the methods mentioned in Section 3. 

All test code is available in the GitHub repository \url{https://github.com/okason97/Bringing-Balance-to-Hand-Shape-Classification}.

\subsubsection{Stratified data generation}

A balanced generated dataset of 1000 images per class was used on each of these methods, using more than 1000 images per class granted no major improvement as it reached the variability limit for each class of the generators. The model was then tested on a held-out test set for RWTH and HaGRID. For RWTH, a total of 39,000 images were generated to use as generated training data. Increasing 19 times over the regular data size when using generated data. This reduces the imbalance by oversampling the minority classes. On HaGRID, generated data represents a smaller increase over the total amount of samples of 552,992.


\begin{table}[ht!]
    \centering
    \footnotesize
    \begin{tblr}{
        colspec={llll},
        row{1}={font=\bfseries},
        column{1,2}={font=\itshape},
        row{even}={bg=gray!10},
    }
        Source & Method & RWTH & HaGRID \\
        \toprule
        real & pretrain ImageNet                              & 80.62  & \textbf{91.08}            \\
        \hline
        ReACGAN & pretrain                                    & \textbf{85.34} & 90.69    \\
        ReACGAN & regularization                              & 78.30  & 89.72                     \\
        ReACGAN & mixup                                       & 76.76  & 90.92                    \\
        \hline
        ReACGAN filtered & pretrain                           & 84.38  & 91.03                     \\
        ReACGAN filtered & regularization                     & 78.50  & 90.53                     \\
        ReACGAN filtered & mixup                              & 80.91  & 90.58                     \\
        \hline
        SPADE & pretrain                                      & 80.91  & \textbf{91.08}                     \\
        SPADE & regularization                                & 75.31  & 90.81                     \\
        SPADE & mixup                                         & 80.14  & 90.00                     \\
        \bottomrule
    \end{tblr}
    \caption{Comparison of EfficientNet v2 performance using different training methods. 
    The table displays the performance using complete and filtered generated datasets conditioned on labels and pose. The filtered datasets contain the top 30\% of samples with the highest class conditional probabilities. Accuracy is evaluated on a held-out test set.}
    \label{tab:filteredresults}
\end{table}

\begin{table}[ht!]
    \centering
    \footnotesize
    \begin{tblr}{
        colspec={llll},
        row{1}={font=\bfseries},
        column{1}={font=\itshape},
        column{1}={5cm},
        row{even}={bg=gray!10},
    }
        Model                                     & RWTH \\
        \toprule
        EfficientNet v2 [Ours]                    & 80.6 \\
        EfficientNet v2 + pretrain with GAN [Ours]   & \textbf{85.3} \\
        EfficientNet v2 + multi-source pretrain with GAN [Ours] & \textbf{85.2} \\
        \hline
        VGG16 \cite{Quiroga2017}                  & 82.8 \\
        Inception-ResNet-v2 \cite{Rakowski2018}   & 84.3 \\
        Hand SubUNet \cite{Camgoz2017}            & 80.3 \\
        Koller et al. \cite{Koller2016}           & 62.8 \\
        \bottomrule
    \end{tblr}
    \caption{Accuracy of models from multiple sources trained on RWTH. The first two models in the table are the best-performing models we trained. In contrast with the other authors, we do not use further data augmentation or hyperparameter fine-tuning of our model. We also do not use any external data source other than our generated data.}
    \label{tab:resultscomparison}
\end{table}

\begin{table}[ht!]
    \centering
    \footnotesize
    \begin{tblr}{
        colspec={lll},
        row{1}={font=\bfseries},
        column{1,2}={font=\itshape},
        row{even}={bg=gray!10},
    }
        Source & Method & RWTH \\
        \toprule
        RWTH & pretrain ImageNet & 80.62            \\
        \hline
        HaGRID SPADE & multi-source pretrain  & \textbf{85.15} \\
        HaGRID SPADE & multi-source regularization  & 77.43 \\
        HaGRID SPADE & multi-source mixup  & 72.52 \\
        \bottomrule
    \end{tblr}
    \caption{Comparison of EfficientNet v2 performance using different multi-source training methods. Accuracy is evaluated on a held-out test set. Multi-source methods consist of using data generated with a model trained on a different dataset than the objective dataset. In this case, the generator was trained with HaGRID and then used in combination with real data to train a classifier on RWTH.}
    \label{tab:multi-source}
\end{table}

The results presented in Table \ref{tab:filteredresults} indicate that training EfficientNet v2 using generated data improved the performance on RWTH, especially when the generated data was used to pretrain the model. Our model using generated data even outperformed other models trained on the same dataset as displayed in Table \ref{tab:resultscomparison} even considering that our model is trained without the usage of external data sources or further data augmentation. However, on HaGRID, training only with real data proved to be the best option. This is probably due to the large amount of available real labeled data of this dataset. In most cases, using mixup or regularization resulted in similar or lower performance compared to using only real data or pre-training with generated samples. Filtering the top-k samples yielded no noticeable improvement in accuracy. This could be a result of decreasing the amount and variety of the generated samples, which lowers the accuracy increase that would have come from an increase in the overall quality of the images. 

Additionally, to test the viability of multi-task learning with our methods, using our SPADE model trained on HaGRID, we generate a new RWTH-like handshape dataset by feeding RWTH poses to the generator. Then, we ran each of our methods in a multi-source way, fine-tuning with real data from RWTH and generating data from a generator trained on an external source (HaGRID). Results of these experiments can be seen in Table \ref{tab:multi-source}. Multi-source training demonstrated an improvement in RWTH similar to training the same model in a single-source way. Overall, this can extend the applicability of pre-training with generated data to other sign language datasets by reusing the HaGRID SPADE generator.

\subsubsection{Analysis of class accuracy}


\begin{figure}[ht!]
    \centering
    \includegraphics[width=\hsize]{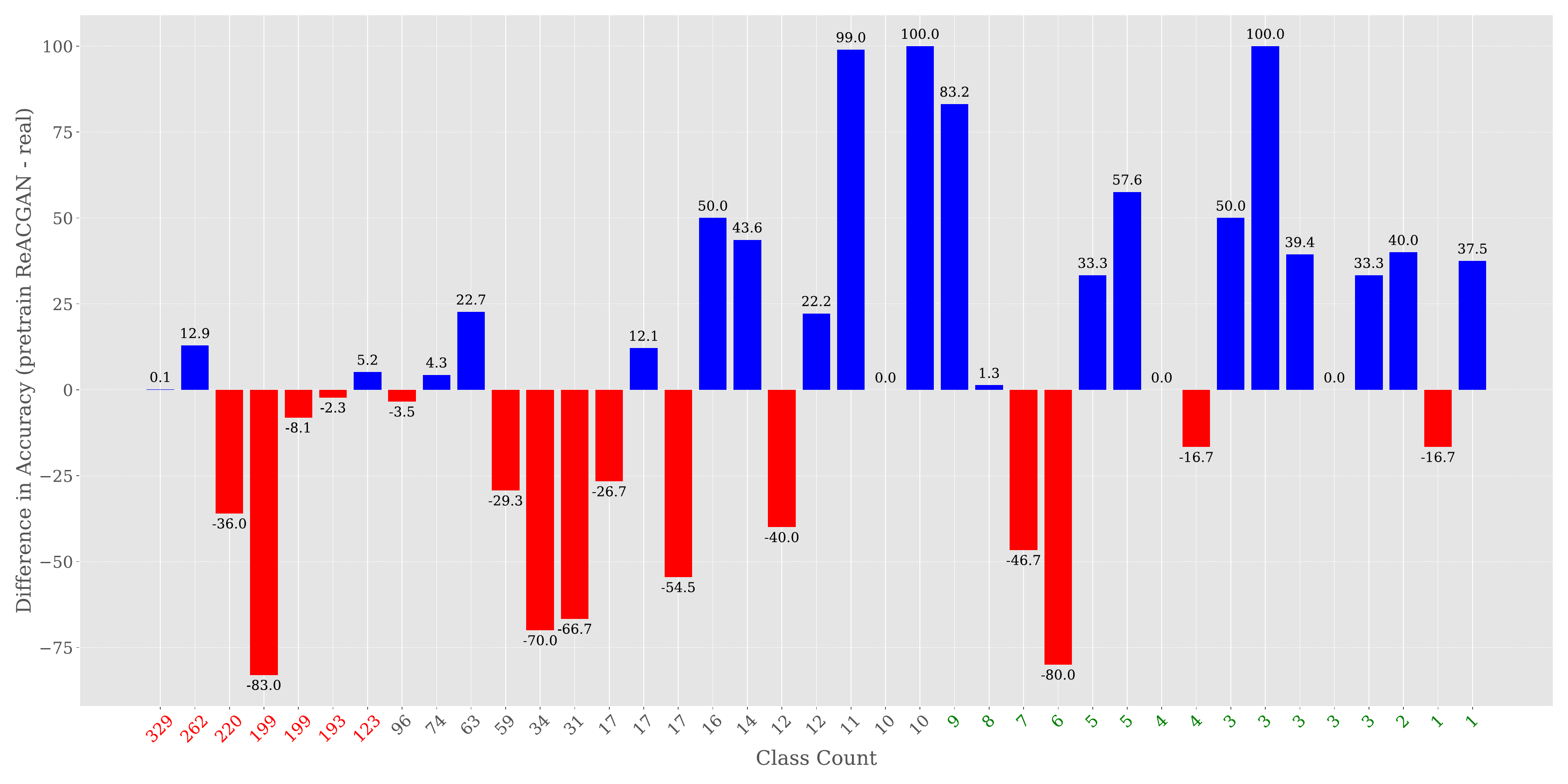}
    \caption{Per-class accuracy difference in RWTH between the ReACGAN-pre-trained model and the baseline model, where each bar represents an individual class. The x-axis shows the number of training samples per class (green for $<$10 samples, red for $>$100 samples), while the y-axis represents the accuracy difference, calculated by subtracting the accuracy of the baseline model from the ReACGAN-pre-trained model. Blue bars indicate improvement with ReACGAN pre-training, red bars show decreased performance. Note the significant improvements for many minority classes with few samples.}
    \label{fig:classaccuracy}
\end{figure}

While using samples generated from a GAN improved the total accuracy of our model, it's important to analyze the per-class performance to understand how this improvement is distributed across majority and minority classes. Figure \ref{fig:classaccuracy} presents a detailed comparison of the per-class accuracy differences in RWTH between our model pre-trained with ReACGAN-generated data and the baseline model trained only on real data. We calculate this difference by subtracting the accuracy of the baseline model from that of the ReACGAN-pre-trained model. A positive difference indicates that the model trained using generated data has a higher accuracy on that specific class, while a negative difference indicates the opposite.

The results demonstrate that pre-training with ReACGAN-generated data effectively addresses the core issue of data imbalance. We observe significant accuracy improvements for many minority classes, with some classes achieving 100\% accuracy where the baseline model completely failed to classify them. This is particularly noteworthy for classes with extremely few samples (1-3 training instances), where the pre-trained model successfully classified instances that the baseline model completely missed. 

Importantly, these improvements in minority class performance do not come at a substantial cost to the accuracy of the model. While some classes with larger sample sizes show decreases in performance, these are generally outweighed by the gains in other classes, resulting in a $\sim5\%$ higher overall accuracy.

This enhanced ability to classify minority classes without overfitting to majority classes is especially remarkable given that we did not apply any other rebalancing techniques such as class weighting, oversampling, or undersampling. The ReACGAN pre-training approach alone was sufficient to significantly mitigate the effects of class imbalance, demonstrating its effectiveness as a data augmentation strategy for imbalanced datasets.

\subsubsection{Convergence speed evaluation}

\begin{figure}[ht!]
    \begin{subfigure}{0.45\textwidth}
        \includegraphics[width=\hsize]{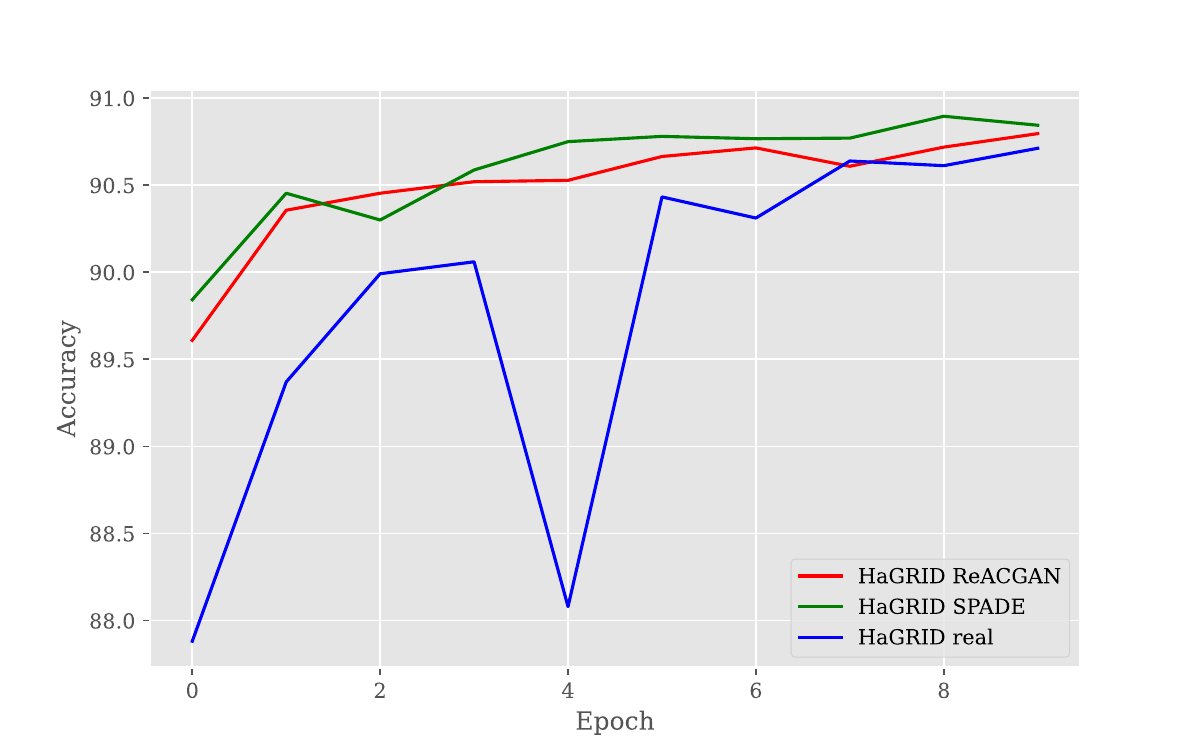}
        \caption{HaGRID}
        \label{fig:HaGRID-reacgan-training}
    \end{subfigure}
    \hfill
    \begin{subfigure}{0.45\textwidth}
        \includegraphics[width=\hsize]{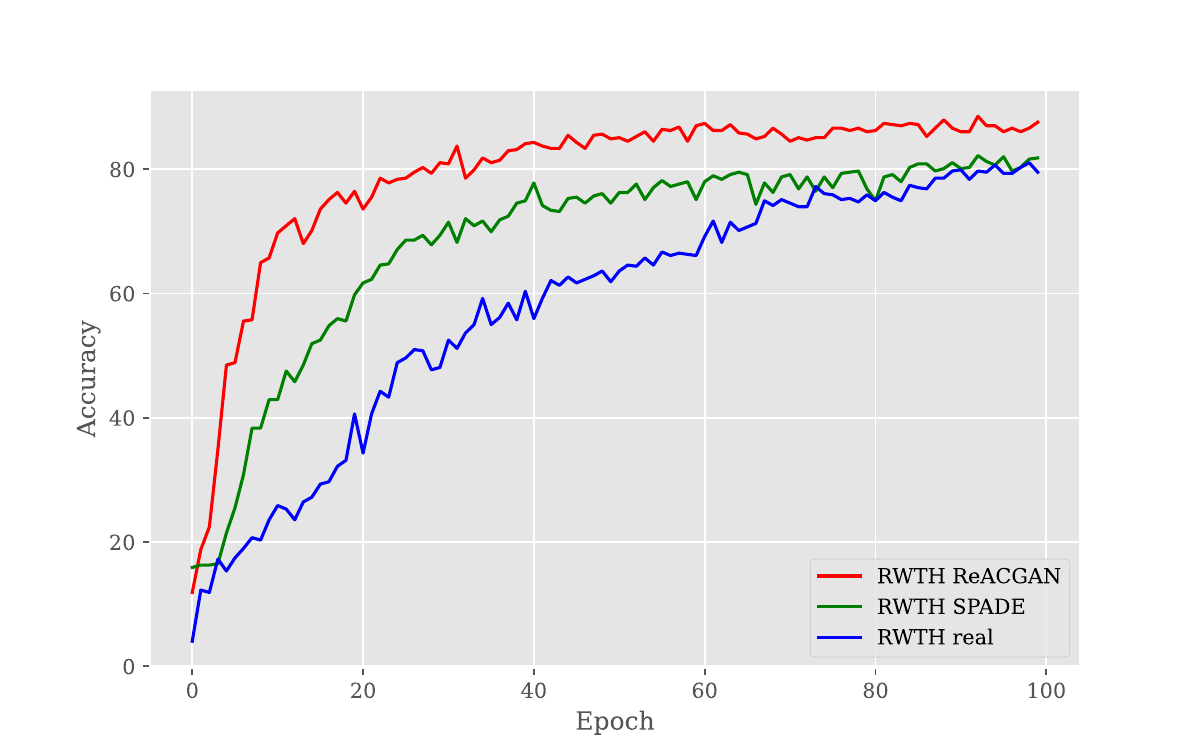}
        \caption{RWTH}
        \label{fig:rwth-spade-training}
    \end{subfigure}
    \caption{Plots showing the accuracy of the classifier model on the training dataset in each epoch. Each plot displays the training of the model trained on HaGRID (left) and RWTH (right). The red line shows the training accuracy per epoch of the model pre-trained with data created using ReACGAN, the green line displays the model trained with data generated by SPADE and the blue line displays the model trained only with real data.}
    \label{fig:training-plot}
\end{figure}


We evaluated the convergence time of models pre-trained with generated data. Our models were able to converge much faster after pre-training using generated samples, as shown in Figure \ref{fig:training-plot}. When training with RWTH we were able to train the model in 60\% of the required epochs using only real data. In all cases, pre-training with generated data achieved convergence in about half the epochs required without using this technique. This implies that a domain-specific initialization using generated data can be a good way to approximate global minima at the start of the training.

\subsubsection{Handshape classification with limited data}

\begin{table}[ht!]
    \centering
    \footnotesize
    \begin{tblr}{
        colspec={lllll},
        row{1}={font=\bfseries},
        column{1-2}={font=\itshape},
        column{1-2}={3cm},
        row{even}={bg=gray!10},
    }
        Source & Method & 5 & 10 & 20 & all\\
        \toprule
        RWTH         & pretrain Imagenet & 1.89 & 4.11 & 53.16 & 80.62 \\
        \hline
        RWTH ReACGAN & pretrain          & \textbf{37.74} & \textbf{54.16} & \textbf{75.58} & \textbf{85.34} \\
        RWTH ReACGAN & mixup             & 0.44 & 4.33 & 35.52 & 76.76 \\
        RWTH ReACGAN & regularization    & 15.54 & 25.97 & 44.51 & 78.30 \\
        \hline
        RWTH SPADE   & pretrain          & 13.43 & 17.54 & 73.25 & 80.91 \\
        RWTH SPADE   & mixup             & 0.01 & 1.44 & 54.61 & 80.14 \\
        RWTH SPADE   & regularization    & 31.30 & 37.85 & 34.18 & 75.31 \\
        \hline
        HaGRID SPADE & multi-source pretrain           & 23.31 & 53.39 & 74.81 & 85.15 \\
        HaGRID SPADE & multi-source mixup              & 2.22 & 31.96 & 43.73 & 77.43 \\
        HaGRID SPADE & multi-source regularization     & 23.42 & 14.54 & 39.51 & 72.52\\
        \bottomrule
    \end{tblr}
    \caption{Comparison of model performance on RWTH dataset using different training methods and number of samples per class (5, 10, and 20). The table displays the accuracy scores of an EfficientNet v2 model trained on data generated by GAN models conditioned on labels (ReACGAN) and hand poses (SPADE). Accuracy is evaluated on the same held-out test set for all training set sizes.}
    \label{tab:RWTH classifiers}
\end{table}

\begin{table}[ht!]
    \centering
    \footnotesize
    \begin{tblr}{
        colspec={llllll},
        row{1}={font=\bfseries},
        column{1-2}={font=\itshape},
        column{1-2}={2.93cm},
        row{even}={bg=gray!10},
    }
        Source & Method & 10 & 20 & 40 & all \\
        \toprule
        HaGRID           & pretrain Imagenet  & 9.19 & 40.42 & 68.19 & 91.08  \\
        \hline
        HaGRID ReACGAN   & pretrain           & 47.69 & 70.19 & 80.86 & 90.69 \\
        HaGRID ReACGAN   & mixup              & 13.08 & 25.53 & 70.61 & 90.92 \\
        HaGRID ReACGAN   & regularization     & 28.78 & 48.25 & 60.44 & 89.72 \\
        \hline
        HaGRID SPADE     & pretrain           & \textbf{53.44} & \textbf{74.92} & \textbf{81.92} & \textbf{91.08} \\
        HaGRID SPADE     & mixup              & 13.25 & 33.11 & 70.44 & 90.81 \\
        HaGRID SPADE     & regularization     & 23.61 & 54.22 & 68.53 & 90.00 \\
        \bottomrule
    \end{tblr}
    \caption{Comparison of model performance on HaGRID dataset using different training methods and number of samples per class (10, 20, and 40 samples). The table displays the accuracy scores of an EfficientNet v2 model trained on data generated by GAN models conditioned on labels (ReACGAN) and hand poses (SPADE). Accuracy is evaluated on the same held-out test set for all training set sizes.}
    \label{tab:HaGRID classifiers}
\end{table}

To discover the impact of using generated data to train classifier models on datasets with limited data we ran experiments using reduced variants of each real dataset. This experiment intends to demonstrate the effectiveness of our model on smaller sign language datasets than RWTH. These reduced datasets contain a fixed number of samples per class taken from the original training samples. Due to the difference in complexity of the datasets we took 5, 10, and 20 samples per class for RWTH and 10, 20, and 40 for HaGRID. The testing set remains the same as used for the complete dataset experiments. We trained new generators using the limited datasets as training data and used the new generators to create generated datasets for each real dataset. Then, we trained EfficientNet v2 M on each dataset using our methods of combining generated and real data. 

As shown in tables \ref{tab:RWTH classifiers} and \ref{tab:HaGRID classifiers}, we can see that using generated data significantly improved the performance of the classifier trained with the reduced versions of RWTH and HaGRID. The difference in accuracy is greater when there is less available real data. pre-training with data generated by ReACGAN proved to be the best method of using generated data, achieving the highest accuracy in both RWTH and HaGRID.

\section{Conclusions \& Future Work}
In this article, we propose using ReACGAN and SPADE to generate realistic hand images based on label and pose conditioning, respectively. The models were used to generate balanced datasets that improved the performance of classifiers on the highly imbalanced RWTH handshape dataset.

We measured the realism of the generator models using multiple qualitative and quantitative metrics. Evaluation by human subjects indicates that the models can generate high-quality handshape images. The models conditioned with labels generated better images than those conditioned with poses. This could be due to the usage of a better discriminator loss, which requires further study.

The performance of the classifier models trained with synthetic data was improved, especially for RWTH and the reduced variants of RWTH and HaGRID. 
We obtained an accuracy on RWTH of 85.3\%, beating the current state-of-the-art static hand shape classifiers without needing further data augmentation or external data sources. We showed that pre-training using data created by a generator trained on a different domain could also improve the performance, obtaining an accuracy of 85.15\% on RWTH when pre-training using a generator trained with HaGRID. Our model also showed less overfitting when dealing with unbalanced datasets, being capable of predicting classes with fewer samples that had no true positives on the model trained only with real data. 
Of all the proposed methods to take advantage of generated data, pre-training with synthetic data was consistently the best performing. We also observed faster convergence when pre-training with generated data, significantly reducing the time required for fine-tuning. These results indicate that using datasets created by generator models can be a good approach when dealing with small and unbalanced datasets. 

For future work, we will experiment with domain adaptation using image-to-image generator models with pose information. This aims to increase the performance of the generator when creating out-of-domain images of different sign languages, which would let us further delve into the idea of using a single generator that can be used to train classifiers on any sign language.

\section*{CRediT authorship contribution statement}
\textbf{Gaston Gustavo Rios:} Conceptualization, Data curation, Investigation, Methodology, Software, Validation, Visualization, Writing – original draft

\textbf{Pedro Dal Bianco:} Conceptualization, Methodology

\textbf{Franco Ronchetti:} Conceptualization, Formal Analysis, Methodology, Supervision, Writing – review \& editing

\textbf{Facundo Quiroga:} Conceptualization, Formal Analysis, Methodology, Supervision, Writing – review \& editing

\textbf{Oscar Stanchi:} Methodology, Writing – review \& editing

\textbf{Santiago Ponte Ahón:} Methodology, Writing – review \& editing

\textbf{Waldo Hasperué:} Project administration, Resources

\section*{Declaration of competing interest}
The authors declare that they have no known competing financial interests or personal relationships that could have appeared to influence the work reported in this paper.

\section*{Data availability}
All our research data comes from public datasets. Synthetic datasets can be generated by using the specified models.








\bibliographystyle{elsarticle-harv} 
\bibliography{bibliography.bib}

\begin{thebibliography}{58}
\expandafter\ifx\csname natexlab\endcsname\relax\def\natexlab#1{#1}\fi
\providecommand{\url}[1]{\texttt{#1}}
\providecommand{\href}[2]{#2}
\providecommand{\path}[1]{#1}
\providecommand{\DOIprefix}{doi:}
\providecommand{\ArXivprefix}{arXiv:}
\providecommand{\URLprefix}{URL: }
\providecommand{\Pubmedprefix}{pmid:}
\providecommand{\doi}[1]{\href{http://dx.doi.org/#1}{\path{#1}}}
\providecommand{\Pubmed}[1]{\href{pmid:#1}{\path{#1}}}
\providecommand{\bibinfo}[2]{#2}
\ifx\xfnm\relax \def\xfnm[#1]{\unskip,\space#1}\fi
\bibitem[{Arora and Zhang(2017)}]{Arora2017}
\bibinfo{author}{Arora, S.}, \bibinfo{author}{Zhang, Y.}, \bibinfo{year}{2017}.
\newblock \bibinfo{title}{Do gans actually learn the distribution? an empirical
  study}.
\newblock \bibinfo{journal}{CoRR} \bibinfo{volume}{abs/1706.08224}.
\bibitem[{Baltatzis et~al.(2024)Baltatzis, Potamias, Ververas, Sun, Deng and
  Zafeiriou}]{Baltatzis2024}
\bibinfo{author}{Baltatzis, V.}, \bibinfo{author}{Potamias, R.A.},
  \bibinfo{author}{Ververas, E.}, \bibinfo{author}{Sun, G.},
  \bibinfo{author}{Deng, J.}, \bibinfo{author}{Zafeiriou, S.},
  \bibinfo{year}{2024}.
\newblock \bibinfo{title}{Neural sign actors: A diffusion model for 3d sign
  language production from text}, in: \bibinfo{booktitle}{Proceedings of the
  IEEE/CVF Conference on Computer Vision and Pattern Recognition (CVPR)}, pp.
  \bibinfo{pages}{1985--1995}.
\bibitem[{Behl et~al.(2020)Behl, Baydin, Gal, Torr and Vineet}]{Harkirat2020}
\bibinfo{author}{Behl, H.S.}, \bibinfo{author}{Baydin, A.G.},
  \bibinfo{author}{Gal, R.}, \bibinfo{author}{Torr, P.H.S.},
  \bibinfo{author}{Vineet, V.}, \bibinfo{year}{2020}.
\newblock \bibinfo{title}{Autosimulate: (quickly) learning synthetic data
  generation}, in: \bibinfo{booktitle}{Computer Vision -- ECCV 2020}, pp.
  \bibinfo{pages}{255--271}.
\bibitem[{Betker et~al.(2023)Betker, Goh, Jing, TimBrooks, Wang, Li,
  LongOuyang, JuntangZhuang, JoyceLee, YufeiGuo, WesamManassra,
  PrafullaDhariwal, CaseyChu, YunxinJiao and Ramesh}]{Betker2023}
\bibinfo{author}{Betker, J.}, \bibinfo{author}{Goh, G.}, \bibinfo{author}{Jing,
  L.}, \bibinfo{author}{TimBrooks, â.}, \bibinfo{author}{Wang, J.},
  \bibinfo{author}{Li, L.}, \bibinfo{author}{LongOuyang, â.},
  \bibinfo{author}{JuntangZhuang, â.}, \bibinfo{author}{JoyceLee, â.},
  \bibinfo{author}{YufeiGuo, â.}, \bibinfo{author}{WesamManassra, â.},
  \bibinfo{author}{PrafullaDhariwal, â.}, \bibinfo{author}{CaseyChu, â.},
  \bibinfo{author}{YunxinJiao, â.}, \bibinfo{author}{Ramesh, A.},
  \bibinfo{year}{2023}.
\newblock \bibinfo{title}{Improving image generation with better captions}.
\bibitem[{Bhattarai et~al.(2020)Bhattarai, Baek, Bodur and Kim}]{Binod2019}
\bibinfo{author}{Bhattarai, B.}, \bibinfo{author}{Baek, S.},
  \bibinfo{author}{Bodur, R.}, \bibinfo{author}{Kim, T.K.},
  \bibinfo{year}{2020}.
\newblock \bibinfo{title}{Sampling strategies for {GAN} synthetic data}, in:
  \bibinfo{booktitle}{ICASSP 2020-2020 IEEE International Conference on
  Acoustics, Speech and Signal Processing (ICASSP)}, pp.
  \bibinfo{pages}{2303--2307}.
\bibitem[{Bowles et~al.(2018)Bowles, Chen, Guerrero, Bentley, Gunn, Hammers,
  Dickie, del C.~Vald{\'{e}}s~Hern{\'{a}}ndez, Wardlaw and
  Rueckert}]{Bowles2018}
\bibinfo{author}{Bowles, C.}, \bibinfo{author}{Chen, L.},
  \bibinfo{author}{Guerrero, R.}, \bibinfo{author}{Bentley, P.},
  \bibinfo{author}{Gunn, R.N.}, \bibinfo{author}{Hammers, A.},
  \bibinfo{author}{Dickie, D.A.}, \bibinfo{author}{del
  C.~Vald{\'{e}}s~Hern{\'{a}}ndez, M.}, \bibinfo{author}{Wardlaw, J.M.},
  \bibinfo{author}{Rueckert, D.}, \bibinfo{year}{2018}.
\newblock \bibinfo{title}{{GAN} augmentation: Augmenting training data using
  generative adversarial networks}.
\newblock \bibinfo{journal}{CoRR} \bibinfo{volume}{abs/1810.10863}.
\bibitem[{Bragg et~al.(2021)Bragg, Caselli, Hochgesang, Huenerfauth,
  Katz-Hernandez, Koller, Kushalnagar, Vogler and Ladner}]{Bragg2021}
\bibinfo{author}{Bragg, D.}, \bibinfo{author}{Caselli, N.},
  \bibinfo{author}{Hochgesang, J.A.}, \bibinfo{author}{Huenerfauth, M.},
  \bibinfo{author}{Katz-Hernandez, L.}, \bibinfo{author}{Koller, O.},
  \bibinfo{author}{Kushalnagar, R.}, \bibinfo{author}{Vogler, C.},
  \bibinfo{author}{Ladner, R.E.}, \bibinfo{year}{2021}.
\newblock \bibinfo{title}{The fate landscape of sign language ai datasets: An
  interdisciplinary perspective}.
\newblock \bibinfo{journal}{ACM Trans. Access. Comput.} \bibinfo{volume}{14}.
\bibitem[{Bragg et~al.(2019)Bragg, Koller, Bellard, Berke, Boudreault,
  Braffort, Caselli, Huenerfauth, Kacorri, Verhoef et~al.}]{Bragg2019}
\bibinfo{author}{Bragg, D.}, \bibinfo{author}{Koller, O.},
  \bibinfo{author}{Bellard, M.}, \bibinfo{author}{Berke, L.},
  \bibinfo{author}{Boudreault, P.}, \bibinfo{author}{Braffort, A.},
  \bibinfo{author}{Caselli, N.}, \bibinfo{author}{Huenerfauth, M.},
  \bibinfo{author}{Kacorri, H.}, \bibinfo{author}{Verhoef, T.}, et~al.,
  \bibinfo{year}{2019}.
\newblock \bibinfo{title}{Sign language recognition, generation, and
  translation: An interdisciplinary perspective}, in:
  \bibinfo{booktitle}{Proceedings of the 21st International ACM SIGACCESS
  Conference on Computers and Accessibility}, pp. \bibinfo{pages}{16--31}.
\bibitem[{Camgoz et~al.(2017)Camgoz, Hadfield, Koller and Bowden}]{Camgoz2017}
\bibinfo{author}{Camgoz, N.C.}, \bibinfo{author}{Hadfield, S.},
  \bibinfo{author}{Koller, O.}, \bibinfo{author}{Bowden, R.},
  \bibinfo{year}{2017}.
\newblock \bibinfo{title}{Subunets: End-to-end hand shape and continuous sign
  language recognition}, in: \bibinfo{booktitle}{2017 IEEE International
  Conference on Computer Vision (ICCV)}, pp. \bibinfo{pages}{3075--3084}.
\bibitem[{Cao et~al.(2023)Cao, Tan, Gao, Xu, Chen, Heng and Li}]{Cao2023}
\bibinfo{author}{Cao, H.}, \bibinfo{author}{Tan, C.}, \bibinfo{author}{Gao,
  Z.}, \bibinfo{author}{Xu, Y.}, \bibinfo{author}{Chen, G.},
  \bibinfo{author}{Heng, P.A.}, \bibinfo{author}{Li, S.Z.},
  \bibinfo{year}{2023}.
\newblock \bibinfo{title}{A survey on generative diffusion model}.
\newblock \href{http://arxiv.org/abs/2209.02646}{{\tt arXiv:2209.02646}}.
\bibitem[{Cao et~al.(2021)Cao, Hidalgo, Simon, Wei and Sheikh}]{Cao2019}
\bibinfo{author}{Cao, Z.}, \bibinfo{author}{Hidalgo, G.},
  \bibinfo{author}{Simon, T.}, \bibinfo{author}{Wei, S.E.},
  \bibinfo{author}{Sheikh, Y.}, \bibinfo{year}{2021}.
\newblock \bibinfo{title}{{ OpenPose: Realtime Multi-Person 2D Pose Estimation
  Using Part Affinity Fields }}.
\newblock \bibinfo{journal}{IEEE Transactions on Pattern Analysis \& Machine
  Intelligence} \bibinfo{volume}{43}, \bibinfo{pages}{172--186}.
\bibitem[{Cornejo~Fandos et~al.(2019)Cornejo~Fandos, Rios, Ronchetti, Quiroga,
  Hasperu{\'e} and Lanzarini}]{cornejo2019}
\bibinfo{author}{Cornejo~Fandos, U.J.}, \bibinfo{author}{Rios, G.G.},
  \bibinfo{author}{Ronchetti, F.}, \bibinfo{author}{Quiroga, F.},
  \bibinfo{author}{Hasperu{\'e}, W.}, \bibinfo{author}{Lanzarini, L.C.},
  \bibinfo{year}{2019}.
\newblock \bibinfo{title}{Recognizing handshapes using small datasets}, in:
  \bibinfo{booktitle}{XXV Congreso Argentino de Ciencias de la Computaci{\'o}n
  (CACIC 2019, Universidad Nacional de Río Cuarto)}.
\bibitem[{Dal~Bianco et~al.(2022)Dal~Bianco, R{\'i}os, Ronchetti, Quiroga,
  Stanchi, Hasperu{\'e} and Rosete}]{DalBianco2022}
\bibinfo{author}{Dal~Bianco, P.}, \bibinfo{author}{R{\'i}os, G.},
  \bibinfo{author}{Ronchetti, F.}, \bibinfo{author}{Quiroga, F.},
  \bibinfo{author}{Stanchi, O.}, \bibinfo{author}{Hasperu{\'e}, W.},
  \bibinfo{author}{Rosete, A.}, \bibinfo{year}{2022}.
\newblock \bibinfo{title}{Lsa-t: The first continuous argentinian sign language
  dataset for sign language translation}, in: \bibinfo{booktitle}{Advances in
  Artificial Intelligence -- IBERAMIA 2022}, pp. \bibinfo{pages}{293--304}.
\bibitem[{Dhariwal and Nichol(2021)}]{Dhariwal2021}
\bibinfo{author}{Dhariwal, P.}, \bibinfo{author}{Nichol, A.},
  \bibinfo{year}{2021}.
\newblock \bibinfo{title}{Diffusion models beat gans on image synthesis}.
\newblock \bibinfo{journal}{Advances in neural information processing systems}
  \bibinfo{volume}{34}, \bibinfo{pages}{8780--8794}.
\bibitem[{Fowley and Ventresque(2021)}]{Fowley2022}
\bibinfo{author}{Fowley, F.}, \bibinfo{author}{Ventresque, A.},
  \bibinfo{year}{2021}.
\newblock \bibinfo{title}{Sign language fingerspelling recognition using
  synthetic data}, in: \bibinfo{booktitle}{Irish Conference on Artificial
  Intelligence and Cognitive Science}.
\bibitem[{Frid-Adar et~al.(2018)Frid-Adar, Klang, Amitai, Goldberger and
  Greenspan}]{Maayan2018}
\bibinfo{author}{Frid-Adar, M.}, \bibinfo{author}{Klang, E.},
  \bibinfo{author}{Amitai, M.}, \bibinfo{author}{Goldberger, J.},
  \bibinfo{author}{Greenspan, H.}, \bibinfo{year}{2018}.
\newblock \bibinfo{title}{Synthetic data augmentation using {GAN} for improved
  liver lesion classification}, in: \bibinfo{booktitle}{2018 IEEE 15th
  international symposium on biomedical imaging (ISBI 2018)}, pp.
  \bibinfo{pages}{289--293}.
\bibitem[{Gaggiotti(2021)}]{Gaggiotti2021}
\bibinfo{author}{Gaggiotti, W.}, \bibinfo{year}{2021}.
\newblock \bibinfo{title}{Redes GANs como t{\'e}cnica de data augmentation para
  el reconocimiento de lengua de senas}.
\newblock Ph.D. thesis. Universidad Nacional de La Plata.
\bibitem[{Hataya and Nakayama(2019)}]{Hataya2019}
\bibinfo{author}{Hataya, R.}, \bibinfo{author}{Nakayama, H.},
  \bibinfo{year}{2019}.
\newblock \bibinfo{title}{Unifying semi-supervised and robust learning by
  mixup} .
\bibitem[{Heusel et~al.(2017)Heusel, Ramsauer, Unterthiner, Nessler and
  Hochreiter}]{Heusel2017}
\bibinfo{author}{Heusel, M.}, \bibinfo{author}{Ramsauer, H.},
  \bibinfo{author}{Unterthiner, T.}, \bibinfo{author}{Nessler, B.},
  \bibinfo{author}{Hochreiter, S.}, \bibinfo{year}{2017}.
\newblock \bibinfo{title}{Gans trained by a two time-scale update rule converge
  to a local nash equilibrium}, in: \bibinfo{booktitle}{Advances in Neural
  Information Processing Systems}.
\bibitem[{Ibrahim and Kashef(2012)}]{Ibrahim2012}
\bibinfo{author}{Ibrahim, A.}, \bibinfo{author}{Kashef, R.},
  \bibinfo{year}{2012}.
\newblock \bibinfo{title}{Visual synthetic data generation for sign language
  recognition}.
\bibitem[{Jiang et~al.(2009)Jiang, Gao, Yao, Zhao and Chen}]{Jiang2009}
\bibinfo{author}{Jiang, F.}, \bibinfo{author}{Gao, W.}, \bibinfo{author}{Yao,
  H.}, \bibinfo{author}{Zhao, D.}, \bibinfo{author}{Chen, X.},
  \bibinfo{year}{2009}.
\newblock \bibinfo{title}{Synthetic data generation technique in
  signer-independent sign language recognition}.
\newblock \bibinfo{journal}{Pattern Recognition Letters} \bibinfo{volume}{30},
  \bibinfo{pages}{513--524}.
\bibitem[{Kang et~al.(2021)Kang, Shim, Cho and Park}]{Kang2021}
\bibinfo{author}{Kang, M.}, \bibinfo{author}{Shim, W.}, \bibinfo{author}{Cho,
  M.}, \bibinfo{author}{Park, J.}, \bibinfo{year}{2021}.
\newblock \bibinfo{title}{Rebooting {ACGAN:} auxiliary classifier gans with
  stable training}.
\newblock \bibinfo{journal}{Advances in neural information processing systems}
  \bibinfo{volume}{34}, \bibinfo{pages}{23505--23518}.
\bibitem[{Kapitanov et~al.(2024)Kapitanov, Kvanchiani, Nagaev, Kraynov and
  Makhliarchuk}]{Kapitanov2022}
\bibinfo{author}{Kapitanov, A.}, \bibinfo{author}{Kvanchiani, K.},
  \bibinfo{author}{Nagaev, A.}, \bibinfo{author}{Kraynov, R.},
  \bibinfo{author}{Makhliarchuk, A.}, \bibinfo{year}{2024}.
\newblock \bibinfo{title}{Hagrid -- hand gesture recognition image dataset},
  in: \bibinfo{booktitle}{Proceedings of the IEEE/CVF Winter Conference on
  Applications of Computer Vision (WACV)}, pp. \bibinfo{pages}{4572--4581}.
\bibitem[{Kim and O{'}Neill-Brown(2019)}]{Kim2019}
\bibinfo{author}{Kim, J.}, \bibinfo{author}{O{'}Neill-Brown, P.},
  \bibinfo{year}{2019}.
\newblock \bibinfo{title}{Improving {A}merican {S}ign {L}anguage recognition
  with synthetic data}, in: \bibinfo{booktitle}{Proceedings of Machine
  Translation Summit XVII: Research Track}, pp. \bibinfo{pages}{151--161}.
\bibitem[{Koller(2020)}]{Koller2020}
\bibinfo{author}{Koller, O.}, \bibinfo{year}{2020}.
\newblock \bibinfo{title}{Quantitative survey of the state of the art in sign
  language recognition}.
\newblock \bibinfo{journal}{CoRR} \bibinfo{volume}{abs/2008.09918}.
\bibitem[{Koller et~al.(2016)Koller, Ney and Bowden}]{Koller2016}
\bibinfo{author}{Koller, O.}, \bibinfo{author}{Ney, H.},
  \bibinfo{author}{Bowden, R.}, \bibinfo{year}{2016}.
\newblock \bibinfo{title}{Deep hand: How to train a cnn on 1 million hand
  images when your data is continuous and weakly labelled}, in:
  \bibinfo{booktitle}{2016 IEEE Conference on Computer Vision and Pattern
  Recognition (CVPR)}, pp. \bibinfo{pages}{3793--3802}.
\bibitem[{Konstantinov and Lampert(2019)}]{Konstantinov2019}
\bibinfo{author}{Konstantinov, N.}, \bibinfo{author}{Lampert, C.},
  \bibinfo{year}{2019}.
\newblock \bibinfo{title}{Robust learning from untrusted sources}, in:
  \bibinfo{booktitle}{International conference on machine learning}, pp.
  \bibinfo{pages}{3488--3498}.
\bibitem[{Kortylewski et~al.(2018)Kortylewski, Schneider, Gerig, Egger,
  Morel{-}Forster and Vetter}]{Kortylewski2018}
\bibinfo{author}{Kortylewski, A.}, \bibinfo{author}{Schneider, A.},
  \bibinfo{author}{Gerig, T.}, \bibinfo{author}{Egger, B.},
  \bibinfo{author}{Morel{-}Forster, A.}, \bibinfo{author}{Vetter, T.},
  \bibinfo{year}{2018}.
\newblock \bibinfo{title}{Training deep face recognition systems with synthetic
  data}.
\newblock \bibinfo{journal}{CoRR} \bibinfo{volume}{abs/1802.05891}.
\bibitem[{Krizhevsky et~al.(2012)Krizhevsky, Sutskever and
  Hinton}]{Krizhevsky2012}
\bibinfo{author}{Krizhevsky, A.}, \bibinfo{author}{Sutskever, I.},
  \bibinfo{author}{Hinton, G.E.}, \bibinfo{year}{2012}.
\newblock \bibinfo{title}{Imagenet classification with deep convolutional
  neural networks}, in: \bibinfo{booktitle}{Advances in Neural Information
  Processing Systems}.
\bibitem[{Li et~al.(2020)Li, Rodriguez, Yu and Li}]{Li2020}
\bibinfo{author}{Li, D.}, \bibinfo{author}{Rodriguez, C.}, \bibinfo{author}{Yu,
  X.}, \bibinfo{author}{Li, H.}, \bibinfo{year}{2020}.
\newblock \bibinfo{title}{Word-level deep sign language recognition from video:
  A new large-scale dataset and methods comparison}, in:
  \bibinfo{booktitle}{The IEEE Winter Conference on Applications of Computer
  Vision}, pp. \bibinfo{pages}{1459--1469}.
\bibitem[{Moreno-Barea et~al.(2018)Moreno-Barea, Strazzera, Jerez, Urda and
  Franco}]{Moreno-Barea2018}
\bibinfo{author}{Moreno-Barea, F.J.}, \bibinfo{author}{Strazzera, F.},
  \bibinfo{author}{Jerez, J.M.}, \bibinfo{author}{Urda, D.},
  \bibinfo{author}{Franco, L.}, \bibinfo{year}{2018}.
\newblock \bibinfo{title}{Forward noise adjustment scheme for data
  augmentation}, in: \bibinfo{booktitle}{2018 IEEE Symposium Series on
  Computational Intelligence (SSCI)}, pp. \bibinfo{pages}{728--734}.
\bibitem[{Mostofi et~al.(2024)Mostofi, {Behzat Tokdemir} and
  Toğan}]{Mostofi2024}
\bibinfo{author}{Mostofi, F.}, \bibinfo{author}{{Behzat Tokdemir}, O.},
  \bibinfo{author}{Toğan, V.}, \bibinfo{year}{2024}.
\newblock \bibinfo{title}{Generating synthetic data with variational
  autoencoder to address class imbalance of graph attention network prediction
  model for construction management}.
\newblock \bibinfo{journal}{Advanced Engineering Informatics}
  \bibinfo{volume}{62}, \bibinfo{pages}{102606}.
\bibitem[{Müller-Franzes et~al.(2023)Müller-Franzes, Niehues, Khader,
  Arasteh, Haarburger, Kuhl, Wang, Han, Nolte, Nebelung, Kather and
  Truhn}]{MullerFranzes2023}
\bibinfo{author}{Müller-Franzes, G.}, \bibinfo{author}{Niehues, J.M.},
  \bibinfo{author}{Khader, F.}, \bibinfo{author}{Arasteh, S.T.},
  \bibinfo{author}{Haarburger, C.}, \bibinfo{author}{Kuhl, C.},
  \bibinfo{author}{Wang, T.}, \bibinfo{author}{Han, T.},
  \bibinfo{author}{Nolte, T.}, \bibinfo{author}{Nebelung, S.},
  \bibinfo{author}{Kather, J.N.}, \bibinfo{author}{Truhn, D.},
  \bibinfo{year}{2023}.
\newblock \bibinfo{title}{A multimodal comparison of latent denoising diffusion
  probabilistic models and generative adversarial networks for medical image
  synthesis}.
\newblock \bibinfo{journal}{Scientific Reports} \bibinfo{volume}{13}.
\bibitem[{Naeem et~al.(2020)Naeem, Oh, Uh, Choi and Yoo}]{Naeem2020}
\bibinfo{author}{Naeem, M.F.}, \bibinfo{author}{Oh, S.J.}, \bibinfo{author}{Uh,
  Y.}, \bibinfo{author}{Choi, Y.}, \bibinfo{author}{Yoo, J.},
  \bibinfo{year}{2020}.
\newblock \bibinfo{title}{Reliable fidelity and diversity metrics for
  generative models}, in: \bibinfo{booktitle}{Proceedings of the 37th
  International Conference on Machine Learning}.
\bibitem[{Nagarajan et~al.(2018)Nagarajan, Raffel and
  Goodfellow}]{Vaishnavh2019}
\bibinfo{author}{Nagarajan, V.}, \bibinfo{author}{Raffel, C.},
  \bibinfo{author}{Goodfellow, I.J.}, \bibinfo{year}{2018}.
\newblock \bibinfo{title}{Theoretical insights into memorization in gans}, in:
  \bibinfo{booktitle}{Neural Information Processing Systems Workshop},
  p.~\bibinfo{pages}{3}.
\bibitem[{Núñez-Marcos et~al.(2023)Núñez-Marcos, de~Viñaspre and
  Labaka}]{Nunez2023}
\bibinfo{author}{Núñez-Marcos, A.}, \bibinfo{author}{de~Viñaspre, O.P.},
  \bibinfo{author}{Labaka, G.}, \bibinfo{year}{2023}.
\newblock \bibinfo{title}{A survey on sign language machine translation}.
\newblock \bibinfo{journal}{Expert Systems with Applications}
  \bibinfo{volume}{213}, \bibinfo{pages}{118993}.
\bibitem[{{Odena} et~al.(2016){Odena}, {Olah} and {Shlens}}]{Odena2016}
\bibinfo{author}{{Odena}, A.}, \bibinfo{author}{{Olah}, C.},
  \bibinfo{author}{{Shlens}, J.}, \bibinfo{year}{2016}.
\newblock \bibinfo{title}{{Conditional Image Synthesis With Auxiliary
  Classifier GANs}}.
\newblock \bibinfo{journal}{arXiv e-prints} ,
  \bibinfo{pages}{arXiv:1610.09585}.
\bibitem[{Park et~al.(2019)Park, Liu, Wang and Zhu}]{Park2019}
\bibinfo{author}{Park, T.}, \bibinfo{author}{Liu, M.Y.}, \bibinfo{author}{Wang,
  T.C.}, \bibinfo{author}{Zhu, J.Y.}, \bibinfo{year}{2019}.
\newblock \bibinfo{title}{Semantic image synthesis with spatially-adaptive
  normalization}, in: \bibinfo{booktitle}{Proceedings of the IEEE/CVF
  conference on computer vision and pattern recognition}, pp.
  \bibinfo{pages}{2337--2346}.
\bibitem[{Quiroga et~al.(2017)Quiroga, Antonio, Ronchetti, Lanzarini and
  Rosete}]{Quiroga2017}
\bibinfo{author}{Quiroga, F.}, \bibinfo{author}{Antonio, R.},
  \bibinfo{author}{Ronchetti, F.}, \bibinfo{author}{Lanzarini, L.C.},
  \bibinfo{author}{Rosete, A.}, \bibinfo{year}{2017}.
\newblock \bibinfo{title}{A study of convolutional architectures for handshape
  recognition applied to sign language}, in: \bibinfo{booktitle}{XXIII Congreso
  Argentino de Ciencias de la Computaci{\'o}n (La Plata, 2017).}
\bibitem[{Rakowski and Wandzik(2018)}]{Rakowski2018}
\bibinfo{author}{Rakowski, A.}, \bibinfo{author}{Wandzik, L.},
  \bibinfo{year}{2018}.
\newblock \bibinfo{title}{Hand shape recognition using very deep convolutional
  neural networks}, in: \bibinfo{booktitle}{Proceedings of the 1st
  International Conference on Control and Computer Vision}, p.
  \bibinfo{pages}{8–12}.
\bibitem[{Ren et~al.(2021)Ren, Xiao, Chang, Huang, Li, Gupta, Chen and
  Wang}]{Pengzhen2020}
\bibinfo{author}{Ren, P.}, \bibinfo{author}{Xiao, Y.}, \bibinfo{author}{Chang,
  X.}, \bibinfo{author}{Huang, P.Y.}, \bibinfo{author}{Li, Z.},
  \bibinfo{author}{Gupta, B.B.}, \bibinfo{author}{Chen, X.},
  \bibinfo{author}{Wang, X.}, \bibinfo{year}{2021}.
\newblock \bibinfo{title}{A survey of deep active learning}.
\newblock \bibinfo{journal}{ACM computing surveys (CSUR)} \bibinfo{volume}{54},
  \bibinfo{pages}{1--40}.
\bibitem[{Rezvani and Wang(2023)}]{Rezvani2023}
\bibinfo{author}{Rezvani, S.}, \bibinfo{author}{Wang, X.},
  \bibinfo{year}{2023}.
\newblock \bibinfo{title}{A broad review on class imbalance learning
  techniques}.
\newblock \bibinfo{journal}{Applied Soft Computing} \bibinfo{volume}{143},
  \bibinfo{pages}{110415}.
\bibitem[{Rombach et~al.(2021)Rombach, Blattmann, Lorenz, Esser and
  Ommer}]{Rombach2021}
\bibinfo{author}{Rombach, R.}, \bibinfo{author}{Blattmann, A.},
  \bibinfo{author}{Lorenz, D.}, \bibinfo{author}{Esser, P.},
  \bibinfo{author}{Ommer, B.}, \bibinfo{year}{2021}.
\newblock \bibinfo{title}{High-resolution image synthesis with latent diffusion
  models}.
\newblock \href{http://arxiv.org/abs/2112.10752}{{\tt arXiv:2112.10752}}.
\bibitem[{Salimans et~al.(2016)Salimans, Goodfellow, Zaremba, Cheung, Radford
  and Chen}]{Salimans2016}
\bibinfo{author}{Salimans, T.}, \bibinfo{author}{Goodfellow, I.},
  \bibinfo{author}{Zaremba, W.}, \bibinfo{author}{Cheung, V.},
  \bibinfo{author}{Radford, A.}, \bibinfo{author}{Chen, X.},
  \bibinfo{year}{2016}.
\newblock \bibinfo{title}{Improved techniques for training gans}, in:
  \bibinfo{booktitle}{Proceedings of the 30th International Conference on
  Neural Information Processing Systems}, p. \bibinfo{pages}{2234–2242}.
\bibitem[{Sampath et~al.(2021)Sampath, Maurtua, Aguilar~Mart{\'i}n and
  Gutierrez}]{Sampath2021}
\bibinfo{author}{Sampath, V.}, \bibinfo{author}{Maurtua, I.},
  \bibinfo{author}{Aguilar~Mart{\'i}n, J.J.}, \bibinfo{author}{Gutierrez, A.},
  \bibinfo{year}{2021}.
\newblock \bibinfo{title}{A survey on generative adversarial networks for
  imbalance problems in computer vision tasks}.
\newblock \bibinfo{journal}{Journal of Big Data} \bibinfo{volume}{8},
  \bibinfo{pages}{27}.
\bibitem[{Shrivastava et~al.(2017)Shrivastava, Pfister, Tuzel, Susskind, Wang
  and Webb}]{Shrivastava2016}
\bibinfo{author}{Shrivastava, A.}, \bibinfo{author}{Pfister, T.},
  \bibinfo{author}{Tuzel, O.}, \bibinfo{author}{Susskind, J.},
  \bibinfo{author}{Wang, W.}, \bibinfo{author}{Webb, R.}, \bibinfo{year}{2017}.
\newblock \bibinfo{title}{Learning from simulated and unsupervised images
  through adversarial training}, in: \bibinfo{booktitle}{Proceedings of the
  IEEE conference on computer vision and pattern recognition}, pp.
  \bibinfo{pages}{2107--2116}.
\bibitem[{Tan and Le(2021)}]{Mingxing2021}
\bibinfo{author}{Tan, M.}, \bibinfo{author}{Le, Q.}, \bibinfo{year}{2021}.
\newblock \bibinfo{title}{Efficientnetv2: Smaller models and faster training},
  in: \bibinfo{booktitle}{International conference on machine learning}, pp.
  \bibinfo{pages}{10096--10106}.
\bibitem[{Tremblay et~al.(2018)Tremblay, Prakash, Acuna, Brophy, Jampani, Anil,
  To, Cameracci, Boochoon and Birchfield}]{Tremblay2018}
\bibinfo{author}{Tremblay, J.}, \bibinfo{author}{Prakash, A.},
  \bibinfo{author}{Acuna, D.}, \bibinfo{author}{Brophy, M.},
  \bibinfo{author}{Jampani, V.}, \bibinfo{author}{Anil, C.},
  \bibinfo{author}{To, T.}, \bibinfo{author}{Cameracci, E.},
  \bibinfo{author}{Boochoon, S.}, \bibinfo{author}{Birchfield, S.},
  \bibinfo{year}{2018}.
\newblock \bibinfo{title}{Training deep networks with synthetic data: Bridging
  the reality gap by domain randomization}, in: \bibinfo{booktitle}{Proceedings
  of the IEEE conference on computer vision and pattern recognition workshops},
  pp. \bibinfo{pages}{969--977}.
\bibitem[{Vahdat and Kreis(2022)}]{Vahdat2022}
\bibinfo{author}{Vahdat, A.}, \bibinfo{author}{Kreis, K.},
  \bibinfo{year}{2022}.
\newblock \bibinfo{title}{Improving diffusion models as an alternative to
  gans}.
\newblock \bibinfo{journal}{NVIDIA Developer Blog} .
\bibitem[{Xia et~al.(2023)Xia, Xu, Chen, Zhang and Zhang}]{Xia2023}
\bibinfo{author}{Xia, Y.}, \bibinfo{author}{Xu, Y.}, \bibinfo{author}{Chen,
  P.}, \bibinfo{author}{Zhang, J.}, \bibinfo{author}{Zhang, Y.},
  \bibinfo{year}{2023}.
\newblock \bibinfo{title}{Generative adversarial network with transformer
  generator for boosting ecg classification}.
\newblock \bibinfo{journal}{Biomedical Signal Processing and Control}
  \bibinfo{volume}{80}, \bibinfo{pages}{104276}.
\bibitem[{Yu et~al.(2024)Yu, Huang, Cheng and Birdal}]{Yu2024}
\bibinfo{author}{Yu, Z.}, \bibinfo{author}{Huang, S.}, \bibinfo{author}{Cheng,
  Y.}, \bibinfo{author}{Birdal, T.}, \bibinfo{year}{2024}.
\newblock \bibinfo{title}{Signavatars: A large-scale 3d sign language holistic
  motion dataset and benchmark}, in: \bibinfo{booktitle}{Proceedings of the
  European Conference on Computer Vision (ECCV)}, pp. \bibinfo{pages}{1--19}.
\bibitem[{Zhang et~al.(2017)Zhang, Ciss{\'{e}}, Dauphin and
  Lopez{-}Paz}]{ZhangHongyi2017}
\bibinfo{author}{Zhang, H.}, \bibinfo{author}{Ciss{\'{e}}, M.},
  \bibinfo{author}{Dauphin, Y.N.}, \bibinfo{author}{Lopez{-}Paz, D.},
  \bibinfo{year}{2017}.
\newblock \bibinfo{title}{mixup: Beyond empirical risk minimization}.
\newblock \bibinfo{journal}{CoRR} \bibinfo{volume}{abs/1710.09412}.
\bibitem[{Zhang et~al.(2020)Zhang, Deng, Kawaguchi, Ghorbani and
  Zou}]{Zhang2020}
\bibinfo{author}{Zhang, L.}, \bibinfo{author}{Deng, Z.},
  \bibinfo{author}{Kawaguchi, K.}, \bibinfo{author}{Ghorbani, A.},
  \bibinfo{author}{Zou, J.Y.}, \bibinfo{year}{2020}.
\newblock \bibinfo{title}{How does mixup help with robustness and
  generalization?}
\newblock \bibinfo{journal}{CoRR} \bibinfo{volume}{abs/2010.04819}.
\bibitem[{Zhang and Yang(2017)}]{ZhangYu2017}
\bibinfo{author}{Zhang, Y.}, \bibinfo{author}{Yang, Q.}, \bibinfo{year}{2017}.
\newblock \bibinfo{title}{{An overview of multi-task learning}}.
\newblock \bibinfo{journal}{National Science Review} \bibinfo{volume}{5},
  \bibinfo{pages}{30--43}.
\bibitem[{Zhong et~al.(2020)Zhong, Zheng, Kang, Li and Yang}]{Zhong2020}
\bibinfo{author}{Zhong, Z.}, \bibinfo{author}{Zheng, L.},
  \bibinfo{author}{Kang, G.}, \bibinfo{author}{Li, S.}, \bibinfo{author}{Yang,
  Y.}, \bibinfo{year}{2020}.
\newblock \bibinfo{title}{Random erasing data augmentation}.
\newblock \bibinfo{journal}{Proceedings of the AAAI Conference on Artificial
  Intelligence} \bibinfo{volume}{34}, \bibinfo{pages}{13001--13008}.
\bibitem[{Zhou et~al.(2022)Zhou, Liu, Qiao, Xiang and Loy}]{Kaiyang2021}
\bibinfo{author}{Zhou, K.}, \bibinfo{author}{Liu, Z.}, \bibinfo{author}{Qiao,
  Y.}, \bibinfo{author}{Xiang, T.}, \bibinfo{author}{Loy, C.C.},
  \bibinfo{year}{2022}.
\newblock \bibinfo{title}{Domain generalization: A survey}.
\newblock \bibinfo{journal}{IEEE Transactions on Pattern Analysis and Machine
  Intelligence} \bibinfo{volume}{45}, \bibinfo{pages}{4396--4415}.
\bibitem[{Zhu et~al.(2022)Zhu, Pan, {vanden Broucke} and Xiao}]{Zhu2022}
\bibinfo{author}{Zhu, B.}, \bibinfo{author}{Pan, X.}, \bibinfo{author}{{vanden
  Broucke}, S.}, \bibinfo{author}{Xiao, J.}, \bibinfo{year}{2022}.
\newblock \bibinfo{title}{A gan-based hybrid sampling method for imbalanced
  customer classification}.
\newblock \bibinfo{journal}{Information Sciences} \bibinfo{volume}{609},
  \bibinfo{pages}{1397--1411}.
\bibitem[{Zimmermann and Brox(2017)}]{Zimmermann2017}
\bibinfo{author}{Zimmermann, C.}, \bibinfo{author}{Brox, T.},
  \bibinfo{year}{2017}.
\newblock \bibinfo{title}{Learning to estimate 3d hand pose from single {RGB}
  images}, in: \bibinfo{booktitle}{Proceedings of the IEEE international
  conference on computer vision}, pp. \bibinfo{pages}{4903--4911}.

\end{thebibliography}





\end{document}